%% file: main.tex
\documentclass[runningheads]{llncs}
\usepackage{graphicx}
\usepackage{caption}
\usepackage{subcaption}
\usepackage{float}
\usepackage{tikz}
\usepackage{comment}
\usepackage{amsmath,amssymb} % define this before the line numbering.
\usepackage{cite}
\usepackage{lipsum}
\usepackage{hyperref}
\hypersetup{
    colorlinks=true,
    linkcolor=red,
    filecolor=magenta,      
    urlcolor=magenta,
    pdftitle={Overleaf Example},
    pdfpagemode=FullScreen,
    }
    
\newcommand\blfootnote[1]{%
  \begingroup
  \renewcommand\thefootnote{}\footnote{#1}%
  \addtocounter{footnote}{-1}%
  \endgroup
}
\begin{document}
% \linenumbers
\pagestyle{headings}
\mainmatter
\def\ECCVSubNumber{4581}  % Insert your submission number here

\title{Deep Portrait Delighting} % Replace with your title

\begin{comment}
\titlerunning{ECCV-22 submission ID \ECCVSubNumber} 
\authorrunning{ECCV-22 submission ID \ECCVSubNumber} 
\author{Anonymous ECCV submission}
\institute{Paper ID \ECCVSubNumber}
\end{comment}
%******************

% \begin{comment}
\titlerunning{Deep Portrait Delighting}
\author{Joshua Weir* \orcidID{0000-0002-4445-7278}\and
Junhong Zhao* \orcidID{0000-0001-7031-3828}\and
Andrew Chalmers \orcidID{0000-0001-6457-7341}\and
Taehyun Rhee* \orcidID{0000-0002-6150-0637}}
\authorrunning{J. Weir \textit{et al}.}

\institute{Computational Media Innovation Centre, Victoria University of Wellington, New~Zealand \\
\email{\{josh.weir, j.zhao, andrew.chalmers, taehyun.rhee\}@vuw.ac.nz}\\
https://www.wgtn.ac.nz/cmic}
% \end{comment}
%******************
\maketitle
\blfootnote{* Corresponding author}

\newcommand{\etal}{\textit{et al}. }
\newcommand{\ie}{\textit{i}.\textit{e}. }
\newcommand{\eg}{\textit{e}.\textit{g}. }

\newcommand{\beginsupplement}{%
        \setcounter{table}{0}
        \renewcommand{\thetable}{S\arabic{table}}%
        \setcounter{figure}{0}
        \renewcommand{\thefigure}{S\arabic{figure}}%
     }

\begin{abstract}

\input{abstract}
\end{abstract}

\section{Introduction}
\input{introduction}

\section{Related Work}
\input{related_work}

\section{Method}
\input{proposed_method_2}

\section{Results}
\input{experiments}

\section{Applications}
\input{applications}

\section{Conclusion}
\input{Conclusion}

% \bibliography{egbib}

\input{bibliography}
\title{Supplementary Material: Deep Portrait Delighting}
\author{}
\institute{}
\maketitle

\input{supplementary}

% \title{Supplementary Material: Deep Portrait Delighting}
% \author{}
% \institute{}
% \maketitle

% \input{supplementary}
\end{document}

%% file: abstract.tex
\par We present a deep neural network for removing undesirable shading features from an unconstrained portrait image, recovering the underlying texture. Our training scheme incorporates three regularization strategies: masked loss, to emphasize high-frequency shading features; soft-shadow loss, which improves sensitivity to subtle changes in lighting; and shading-offset estimation, to supervise separation of shading and texture. Our method demonstrates improved delighting quality and generalization when compared with the state-of-the-art. We further demonstrate how our delighting method can enhance the performance of light-sensitive computer vision tasks such as face relighting and semantic parsing, allowing them to handle extreme lighting conditions.\newline

\textbf{Keywords:}~uniform~lighting,~texture recovery,~shadow removal,~portrait

%% file: introduction.tex
Image delighting is a form of image manipulation that aims to remove unwanted lighting features from images to recover the underlying texture. Recovering this underlying texture benefits many light sensitive computer vision tasks such as face recognition, parsing and relighting. Delighting has seen much research within these application areas, but they have only addressed the delighting problem implicitly in their pipeline. None of them have a dedicated delighting solution independent from their application. For example, the \emph{relighting} application area has seen rapid progress with deep neural networks able to render convincing non-lambertian shading effects~\cite{hou2021towards, pandey2021total, wang2020single, nestmeyer2020learning, sun2019single, zhang2021neural}. However, the \emph{delighting} phase is often abstracted. As a consequence, shading features present in the input image often propagate into the output as distortions. In extreme cases, they can alter the identity and perceived structure of portraits, affecting face recognition~\cite{fahmy2006effect, beveridge2010quantifying}.

\begin{figure}
    \centering
    \includegraphics[width=\textwidth]{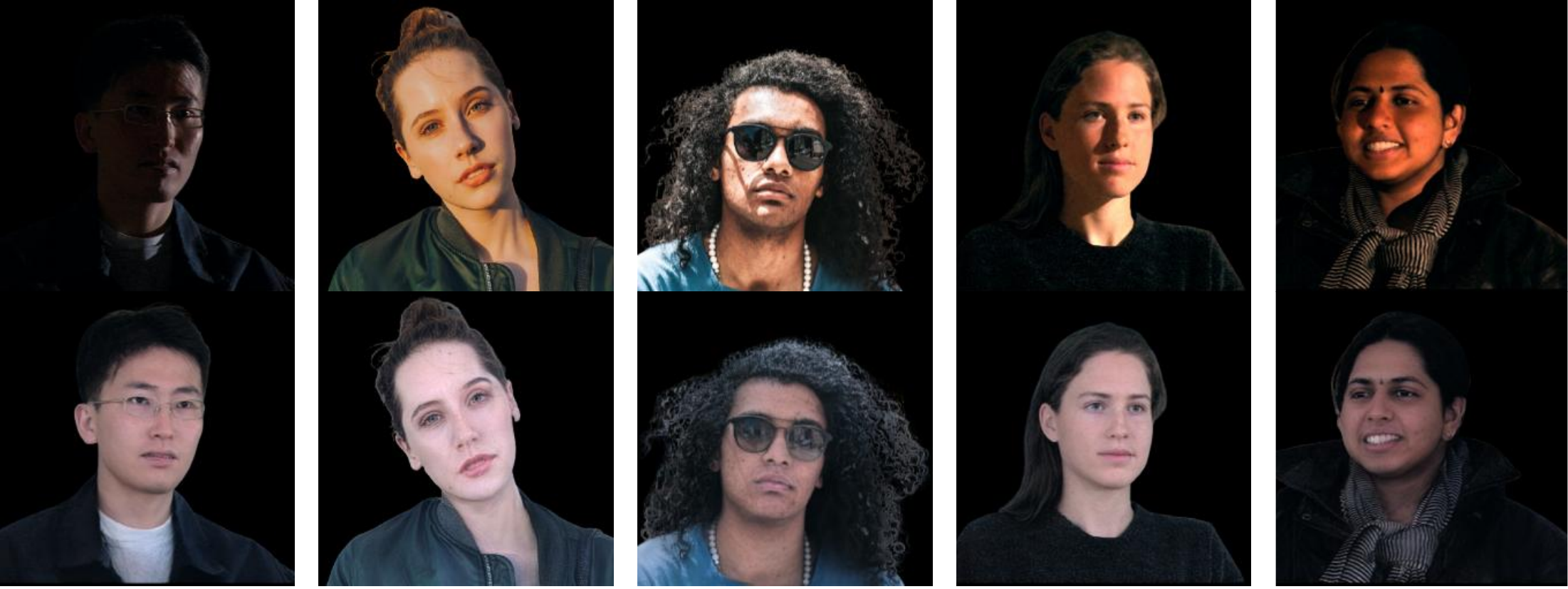}
    \caption{Given a portrait image (top row), we perform \emph{delighting} (bottom row): removing undesirable lighting characteristics and reconstructing the image under uniform lighting.}
    \label{teaser}
\end{figure}

\par The portrait delighting problem is inherently more difficult due to the ambiguous combination of lighting and reflectance determining the colour of any pixel. Theoretically, supervised image-to-image translation pipelines can learn this problem given enough labeled image pairs, but acquiring a representative dataset for this task is laborious and expensive. As a result, most researchers are dependant on 3D renderings of scanned human subjects~\cite{kanamori2018relighting, lagunas2021single, sengupta2018sfsnet}, which consequently leads to poor generalization to real-world images due to oversimplified reflectance and geometry modelling. On the other hand, real human image datasets available to the public suffer from either lighting or subject under-representation. The CMU Multi-PIE dataset~\cite{gross2010multi} for instance, records only 19 lighting conditions for over 300 individuals, while the Extended Yale-B dataset~\cite{KCLee05} provides 64 lightings for only 38 individuals.

\par Even with high-quality data~\cite{sun2019single, pandey2021total}, many challenges still persist, particularly for images exhibiting complex lighting features like reflections and hard shadow boarders; these features have small pixel densities, making common pixel-wise loss functions like $\ell_1$ and $\ell_2$ ineffective; also, these features are highly embedded in the underlying texture we wish to preserve, making them difficult to remove without losing high-frequency details such as freckles and facial hair. Most recent relighting methods that utilize a delighting phase usually incorporate standard network architectures and loss functions which do not directly address the problems of dataset sparsity or non-lambertian lighting.
% \par Perceptual losses have proven to be excellent at preserving both large and small scale details, but require regularization with pixel-wise losses. The optimal loss weights are problem dependent, and in the case of delighting, this creates a trade-off between detail preservation, and the quality of light removal.   

\par We present a fully-supervised portrait delighting method that takes an upper-body portrait lit under an arbitrary illumination, and outputs its reconstruction under uniform white lighting (see Fig.~\ref{teaser}). 

We localize high-frequency lighting effects in our training data using a guided-filter technique, and incorporate these into our training pipeline using a masked loss function inspired by~\cite{hou2021towards} to emphasize small but visually significant regions of the image. 

\par We estimate a shading offset image, which is the difference between the input image and the ground-truth de-lit image. This facilitates learning of lighting features directly, and greatly improves colour consistency. We also synthesize soft-shaded images from our training data and utilize them in training, applying a small regularization loss to their outputs. This alleviates our dataset bias toward directional lightings, allowing us to remove both hard and soft shadows. We demonstrate how our framework benefits the applications of semantic parsing and relighting.
\par To summarize, our main contributions are as follows:
\begin{itemize}
    \item {A novel portrait delighting method that can recover the underlying texture of portraits illuminated under a wide range of complex lighting environments.}
    
    % \item { \sout{A novel portrait delighting method that can serve as a data normalization tool for improving light-sensitive computer-vision tasks; including relighting, face-detection and segmentation.}}
    
    % \item {We assemble a large data}
    
    % \item A shading guided training method for learning the semantic effects of lighting on portrait images. 
    
    % \item A soft-shadow regularization method that improves our models robustness to unseen lighting environments while preserving image detail.
    
    \item {Three novel loss functions: \emph{shading-offset} loss, \emph{soft-shadow} loss and \emph{masked} loss that improve our models robustness to unseen lighting environments while preserving image detail.}
    
    % \item {A masked training method for localizing and removing high-frequency shading features from portraits.}
    
    \item {Our delighting method can serve as a data normalization tool for improving light-sensitive computer vision tasks such as face relighting and semantic parsing.}

\end{itemize}

% \noindent Our delighting method can serve as a data normalization tool for improving light-sensitive computer vision tasks, including, face verification, segmentation and relighting.

%% file: related_work.tex
% \noindent Here, we only cover recent learning-based methods for portrait light-editing, since this subject has a long history in computer vision literature with many target domains. 

% \noindent \textbf{Face Illumination Normalization.}
Much work has been made towards removing disrupting light features (\eg dark shadows, specularities) from faces primarily to enhance face recognition systems~\cite{georghiades2001few, chen2015face}, and to increase image quality~\cite{zhang2020portrait, nagano2019deep, capece2019deepflash}. Pioneering works in this field propose optimization methods using Morphable Face models~\cite{blanz1999morphable, egger20203d, wang2008face, ahmed2007new, chen2013face}. However, relying on parametric models limits their ability to capture non-facial or high-frequency details. Recent deep-learning methods tackle this problem using feature-wise perceptual losses~\cite{ling2020high} or closed-loop GANs~\cite{han2019asymmetric} to recover facial details. A drawback of all these methods however is that they either focus exclusively on face regions, perform only one aspect of the delighting process (\eg shadow removal~\cite{zhang2020portrait}, camera-flash removal~\cite{capece2019deepflash}), or use front-facing illumination as the ground-truth~\cite{nagano2019deep, han2019asymmetric}, ignoring the sharp reflections this causes. 
%  \newline

GAN inversion methods~\cite{abdal2021styleflow, deng2020disentangled, mallikarjun2021photoapp, xia2021gan} enable face editing by projecting images into the latent space of a pre-trained GAN, disentangling lighting, identity, pose and expression attributes such that they can be independently manipulated. These methods can effectively remove sharp shadows and specular reflections, but their reconstructions don't often preserve image content that isn't constrained by editing attributes, particularly high-frequency details such as freckles, and non-face components like clothing.  
 
% \noindent \textbf{Intrinsic Decomposition.}
Intrinsic decomposition methods \emph{delight} by disentangling face~\cite{sengupta2018sfsnet, qiu2020learning, shu2017neural} or full-body~\cite{kanamori2018relighting, lagunas2021single} images into geometry, albedo, and lighting via separate convolutional neural networks, where \emph{relighting} can be performed by re-rendering with modified lighting. Estimation errors usually occur, such as when specularities become embedded in the albedo, or when hard shadows are predicted as geometry features. This becomes a bigger problem when relying on synthetic training data with simplified reflectance models, so Sengupta \etal~\cite{sengupta2018sfsnet} proposed a semi-supervised learning framework using reconstruction loss on real images, but conflation between shading, geometry and texture is still present when there's a significant domain-gap between the real and synthetic data.

Other deep learning methods perform portrait relighting directly~\cite{sun2019single, zhou2019deep, zhang2021neural, hou2021towards, song2021half} by co-opting standard encoder-decoder architectures like U-Net~\cite{ronneberger2015u}. In this manner, both the \emph{delighting} and \emph{relighting} processes are a black box represented by network activations, with supervision applied only to the re-lit image. While this generally leads to more stable and expressive relightings than intrinsic decomposition~\cite{sun2019single}, the neural representation of the delighting task lacks any meaningful form to facilitate supervision. Some researchers attempt to supervise the delighting process by imposing feature-space losses on the network bottleneck~\cite{zhou2019deep, zhang2021neural}. While this helps recover global details such as colour and identity, it often preserves local artifacts such as those caused by facial shadows. 

The most recent relighting methods utilize an explicit \emph{delighting} phase as a preliminary step~\cite{nestmeyer2020learning, wang2020single, pandey2021total}, which is the inspiration for our work. Pandey \etal~\cite{pandey2021total} use a least-squares GAN~\cite{mao2017least} with VGG-perceptual loss~\cite{johnson2016perceptual} in their albedo prediction for accurate detail reconstruction, but colour inconsistencies arise when presented with data unseen in training (\eg different clothing patterns). Wang \etal~\cite{wang2020single} supervise the delighting process by predicting the source lighting and a face-parsing map for more robust texture recovery, but their method struggles to remove high-frequency shading artifacts caused by strong directional light.

%% file: proposed_method_2.tex
\newcommand{\Isrc}{\mathbf{I_{src}}}
\newcommand{\Idlt}{\mathbf{I_{dlt}}}
\newcommand{\Ioff}{\mathbf{I_{off}}}
\newcommand{\Isft}{\mathbf{I_{soft}}}

This section overviews how we synthesize training images (Sec. \ref{dataprocess}), and integrate them into our delighting pipeline (Sec. \ref{basicLoss} to Sec. \ref{high_freq}).

\subsection{Data Processing} \label{dataprocess}

\begin{figure*}
    \begin{center}
        \includegraphics[width=1.0\textwidth]{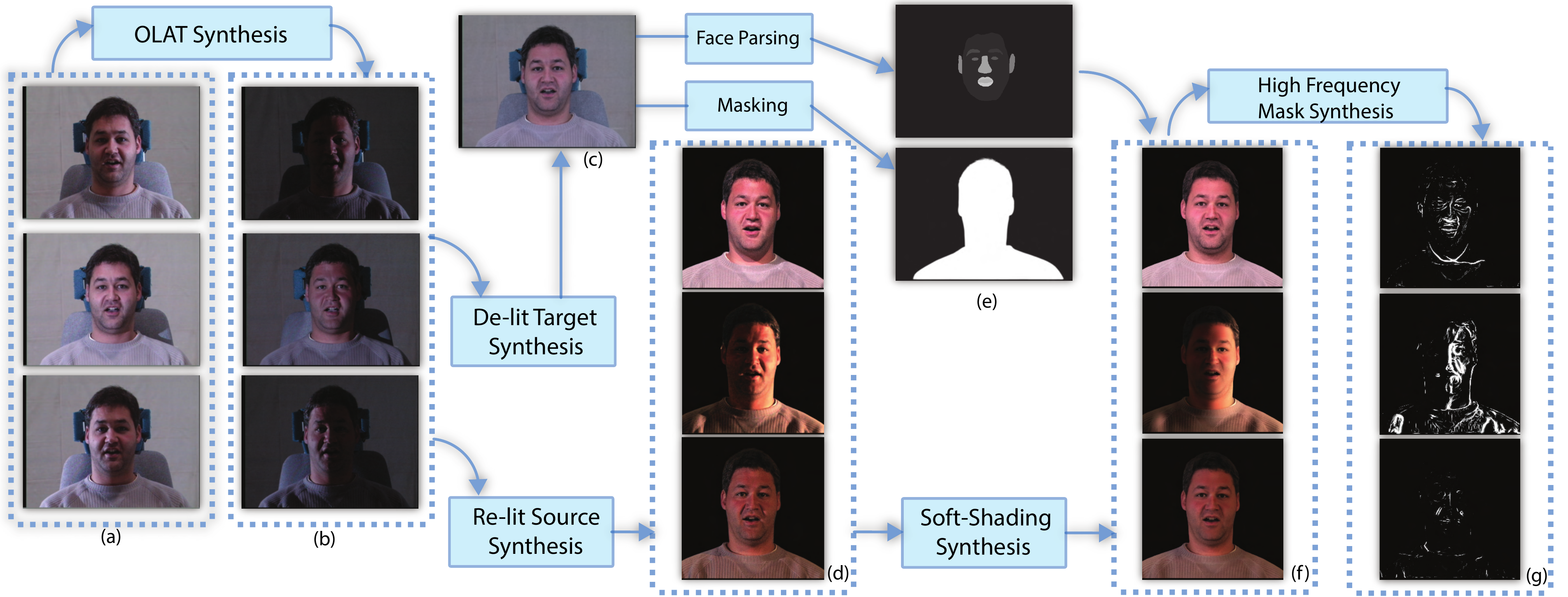}
    \end{center}
    \caption{Our data processing pipeline}
    \label{preprocessing}
\end{figure*}

We modify the Multi-PIE dataset~\cite{gross2010multi} to synthesize our training data for portrait delighting. We chose this dataset as it is the largest publicly available dataset of real subjects. While other datasets capture more lighting directions~\cite{sim2002cmu, georghiades2001few}, they lack diversity in terms of pose, expression, and clothing. Fig.~\ref{preprocessing} illustrates our full data processing pipeline.

% Each static image in the Multi-PIE database was captured under 19 illuminations; 18 directional flashes, and one non-flash image (room lights only). We first remove the effects the room lights have on our flash images (details are described in the supplementary document), resulting in 18 one-light-at-a-time (OLAT) images~\cite{sun2019single} (see Fig.~\ref{preprocessing} (b)), where the target de-lit image ($\Idlt$) is the average over all our OLAT images. From this, we perform foreground masking and face-parsing. (see Fig.~\ref{preprocessing} (e)).

Each static image in the Multi-PIE database was captured under 19 illuminations; 18 directional flashes, and one non-flash image (room lights only). We first remove the effects the room lights have on our flash images (details are described in the supplementary document), resulting in 18 one-light-at-a-time (OLAT) images~\cite{sun2019single} (see Fig.~\ref{preprocessing} (b)), where the target de-lit image ($\Idlt$) is the average over all our OLAT images, with added luminance from our non-flash image. From this, we perform foreground masking and face-parsing. (see Fig.~\ref{preprocessing} (e)).

We stress that the de-lit image $\mathbf{I_{dlt}}$ is different from an albedo image. The difference is that the de-lit image contains subtle light occlusions on non-convex geometric areas such as the ears, nose, and clothing.

% The backgrounds of each subject are removed using commercial software, and semantic labels for each region of the face are extracted using a pre-trained BiSeNet model~\cite{yu2018bisenet} for face parsing. We make use of these segmentation labels to generate soft-shaded images. (see Fig.~\ref{preprocessing} (e)) 

% The original images are at $640\times480$ resolution. We apply an even crop to the left and right sides to produce images at $480\times480$ (1:1 aspect ratio). The target de-lit image ($\mathbf{I_{alb}}$) for our network is the average over all our OLAT images (see Fig.~\ref{preprocessing} (c)). 

To render input images, other learning based methods ~\cite{sun2019single, wang2020single, pandey2021total, zhang2021neural} used Image Based Lighting (IBL). But the mostly front-facing point lights used in Multi-PIE prevent us from adequately sampling 360$^{\circ}$ environment maps. We instead approximate environment illuminations using a weighted average of two OLAT images, each tinted with different colour temperatures (see Fig.~\ref{preprocessing} (d)) similar to the VIDIT dataset ~\cite{helou2020vidit}. This colour assumption in reasonable, as it covers the majority of light found in the real world. High-intensity light images were generated in the same manner by increasing the image brightness, and colour adjusted versions of the de-lit, and room-light-only images were also added to the dataset. 

% This colour assumption in reasonable, as it covers the majority of light found in the real world. Different combinations of direction and colour are used for training and testing

We chose a sample of 140 subjects (70 male, 70 female) under two arbitrary poses, expressions and clothing, producing 240 unique images (220 for training, 20 for testing). We use 1,293 lighting conditions for the training set, and 1,180 for the test set, producing 284,460 images for training, and 25,860 for testing.

\begin{figure*}
    \begin{center}
    % \fbox{\rule{0pt}{2in} \rule{.9\linewidth}{0pt}}
    \includegraphics[clip, trim=0cm 0cm 1.45cm 0cm, width=1.0\textwidth]{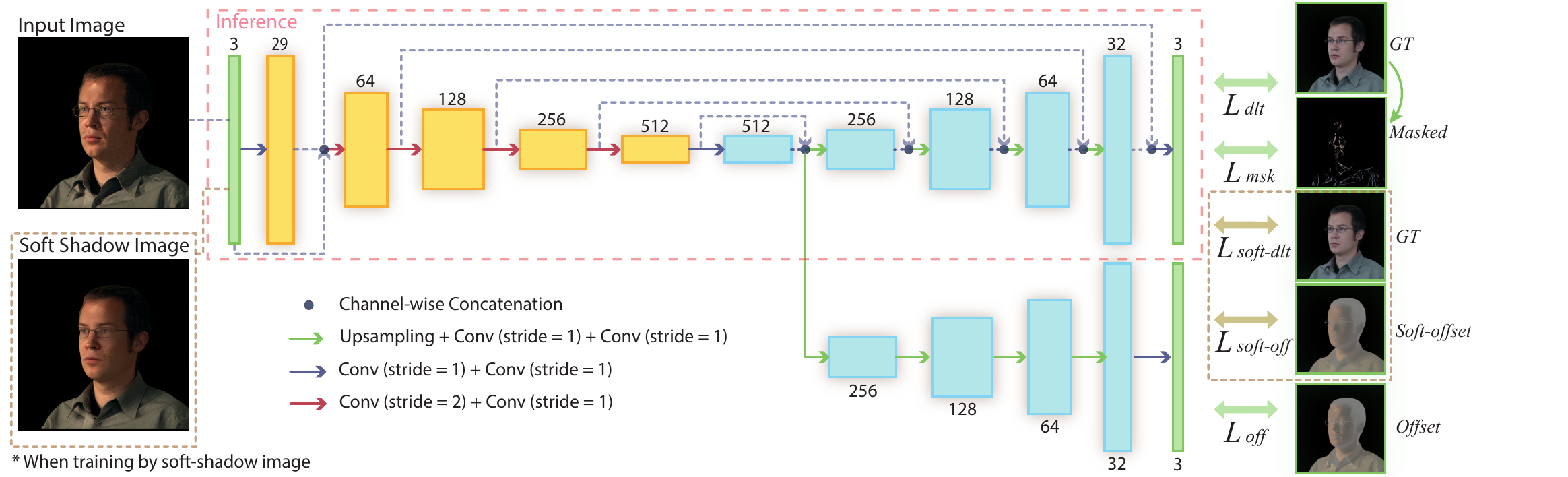}
    \end{center}
    \caption{Overview of our network structure. We predict a shading-offset via a separate decoder during training. All convolutions have a kernel size of 3, and are followed by Instance normalization~\cite{ulyanov2016instance} and PReLU activation~\cite{he2015delving}. \emph{Tanh} activations are used to produce final outputs.}
    \label{fig:architecture}
\end{figure*}

\subsection {Basic Architecture} \label{basicLoss}
We train a U-Net \cite{ronneberger2015u} based CNN to estimate the ground-truth de-lit image $\Idlt$ from a foreground-masked upper-body portrait $\Isrc$ by minimizing a perceptual loss term $\mathcal{L}_\mathbf{perc}$, which based on previous works~\cite{pandey2021total, martin2018lookingood, johnson2016perceptual} has been shown to recover sharp details at multiple scales. This is computed as the $\ell_1$ distance between the activations of the VGG-16 network~\cite{simonyan2014very} pre-trained on ImageNet~\cite{deng2009imagenet}. We also apply a small $\ell_1$ loss to speed up color convergence:

\begin{equation} \label{l_perc}
\begin{split}
    \mathcal{L}_{\mathbf{perc}} (\mathbf{A}, \mathbf{B}) = \sum_{i=1}^{5}\Big[\frac{1}{N_i}&\|VGG_i (\mathbf{A}) - VGG_i(\mathbf{B})\|_{1}\Big]  + \frac{0.2}{M}\|\mathbf{A} - \mathbf{B}\|_ {1},
\end{split}
\end{equation}

\noindent where $\mathbf{A}$ and $\mathbf{B}$ are images of the same subject with $M$ number of foreground pixels, $VGG_i$ and $N_i$ are the outputs and sizes respectively of the final activations before the $i^{th}$ max-pooling layer in VGG-16. We apply this function to our de-lit output $\mathcal{D}_1(\Isrc)$ as:

\begin{equation} \label{dltEq}
    \mathcal{L}_\mathbf{dlt} = \mathcal{L}_{\mathbf{perc}} (\Idlt, \mathcal{D}_1(\Isrc)).
\end{equation}

\noindent This loss alone motivates a direct style transfer from $\Isrc$ to $\Idlt$ without necessarily learning a physical separation of shading and texture, which consequently leads to increased over-fitting to certain modalities of our training data (\ie lighting distribution, clothing patterns) not fully represented in real-world portraits. Shortcomings of this basic architecture are mitigated via our proposed contributions in Sec. 3.3, 3.4 and 3.5

\subsection{Shading Offset}
 We add a separate decoder branch $\mathcal{D}_2$ to our network to learn the difference between $\Isrc$ and $\Idlt$ in the form of a shading-offset image: $\Ioff = \Isrc - \Idlt$. These two images differ only in their shading, so learning the difference between them allows our encoder to learn a meaningful separation of shading and texture. Different from some prior works that obtain their results by subtracting the estimated offset image from the input \cite{capece2019deepflash, nagano2019deep}, we produce our result directly via our delighting decoder $\mathcal{D}_1$ as this makes our model less susceptible to small estimation errors in our shading-offset. Our network pipeline is shown in Fig. \ref{fig:architecture}.

We apply perceptual loss to our offset output $\mathcal{D}_2(\Isrc)$ as:
\begin{equation} \label{offEq}
    \mathcal{L}_\mathbf{off} = \mathcal{L}_{\mathbf{perc}} (\Ioff, \mathcal{D}_2(\Isrc)).
\end{equation}

\subsection{Soft Shadowed Images} \label{soft_shaded_images}
Since the light capture setup of our dataset is too sparse to represent non-directional lighting characteristics such as soft-shadows, we approximate these effects using a   
guided filter~\cite{he2012guided} technique to smooth the source image, while preserving edges in the ground-truth image (see Fig.~\ref{preprocessing} (f)).

\begin{equation}\label{guided_filter}
\begin{split}
    \Isft = &M_{nose} \odot \Omega(\Isrc, \Idlt, \epsilon) + \\
                        &M_{mouth} \odot \Omega(\Isrc, \Idlt, \epsilon) + \\
                        &M_{other} \odot \Omega(\Isrc, \Idlt, \kappa),
\end{split}
\end{equation}

\noindent where $\Omega(I, R, r)$ is the guided filter function ($I$ is the input image, $R$ is the edge reference, and $r$ is the window radius). To preserve morphological features of the face, we use smaller sized filters for the nose and mouth regions ($\epsilon \leq \kappa$). This is done using the masks $M_{nose}$ and $M_{mouth}$ extracted from face parsing. The resulting image approximates the effect of increasing the area of all lights in the input image, softening sharp shadow boarders and specular highlights.

To regularize the strong influence of $\mathcal{L}_\mathbf{dlt}$ and $\mathcal{L}_\mathbf{off}$, we apply a small $\ell_1$ loss to the outputs of $\Isft$ as:

\begin{align}
    &\mathcal{L}_{\mathbf{soft\mbox{-}dlt}} = \frac{0.6}{M}||\Idlt - \mathcal{D}_1(\Isft)||_{1} \label{soft-shading-loss_alb}, \\
    &\mathcal{L}_{\mathbf{soft\mbox{-}off}} = \frac{0.6}{M}||\mathbf{I_{\mathbf{soft\mbox{-}off}}} - \mathcal{D}_2(\Isft)||_{1} \label{soft-shading-loss_off},
\end{align}

\noindent where $\mathbf{I_{\mathbf{soft\mbox{-}off}}} = \Isft - \Idlt$.

\subsection {High-frequency Mask} \label{high_freq}
Similar to \cite{hou2021towards}, We emphasize sharp lighting discontinuities (\eg shadow boarders, reflections) in our loss function via a weight mask $\mathbf{W}$, which we generate using the gradient difference between $\mathbf{I_{src}}$ and its soft-shadowed version (see Fig.~\ref{preprocessing} (g)).

\begin{equation} \label{hf_mask}
\begin{split}
    & a = 10 * \max(\Delta\Isrc - \Delta \Omega(\Isrc, \Idlt, 15), 0) \\
    % & b = \max(a - \Delta\mathbf{I_{alb}}), 0) \\
    & b = median(a) \\
    & \mathbf{W} = \min(b + gauss(b), 1),
\end{split}
\end{equation}

\noindent where $\Delta$ represents the sum of directional gradients along the vertical and horizontal axes. For high-frequency shading features, the gradient of $\mathbf{I_{src}}$ should be higher than its filtered counterpart, which we store in $a$. Afterwards, we apply a median filter to remove noise, and add a small Gaussian blur to increase its receptive field to neighbouring pixels.

Our method is different from Hou \etal~\cite{hou2021towards}, who fit a morphable model to the face, estimate the lighting, and perform ray-casting to produce a shadow mask. Their method is more useful for relighting than delighting, since estimation errors could cause the shadow mask to miss the true shadow boarder. Ours is less vulnerable to estimation errors, and can target areas outside the face.

We compliment the high-level feature activations of $ \mathcal{L}_\mathbf{dlt}$ with this importance mask:

% \begin{equation} \label{l_hf}
% \begin{split}
%     \mathcal{L}_{\mathbf{msk}} =  \sum_{i=1}^{3}\frac{1}{S_i}\|W_i \odot \Big(&VGG_i (\Idlt) - \\
%     &VGG_i(\mathcal{D}_1(\Isrc)\Big)\|_{1},
% \end{split}
% \end{equation}

\begin{equation} \label{l_hf}
    \mathcal{L}_{\mathbf{msk}} =  \sum_{i=1}^{3}\frac{1}{S_i}\|W_i \odot \Big(VGG_i (\Idlt) - VGG_i(\mathcal{D}_1(\Isrc)\Big)\|_{1},
\end{equation}

\noindent where $W_i$ is the high-frequency shading mask resized to fit the dimensions of $VGG_i$, and $S_i$ is the sum of $W_i$.

Our full pipeline is illustrated in Fig. \ref{fig:architecture}, where the final loss is the sum:
\begin{equation}\label{final_loss}
        \mathcal{L} =  \mathcal{L}_{\mathbf{dlt}} + \mathcal{L}_{\mathbf{off}} + \mathcal{L}_{\mathbf{soft\mbox{-}dlt}} + \mathcal{L}_{\mathbf{soft\mbox{-}off}} +  \mathcal{L}_{\mathbf{msk}}.
\end{equation}

%% file: experiments.tex
\newcommand{\imgST}{0.13}

\begin{figure}
    \centering
    \includegraphics[width=\linewidth]{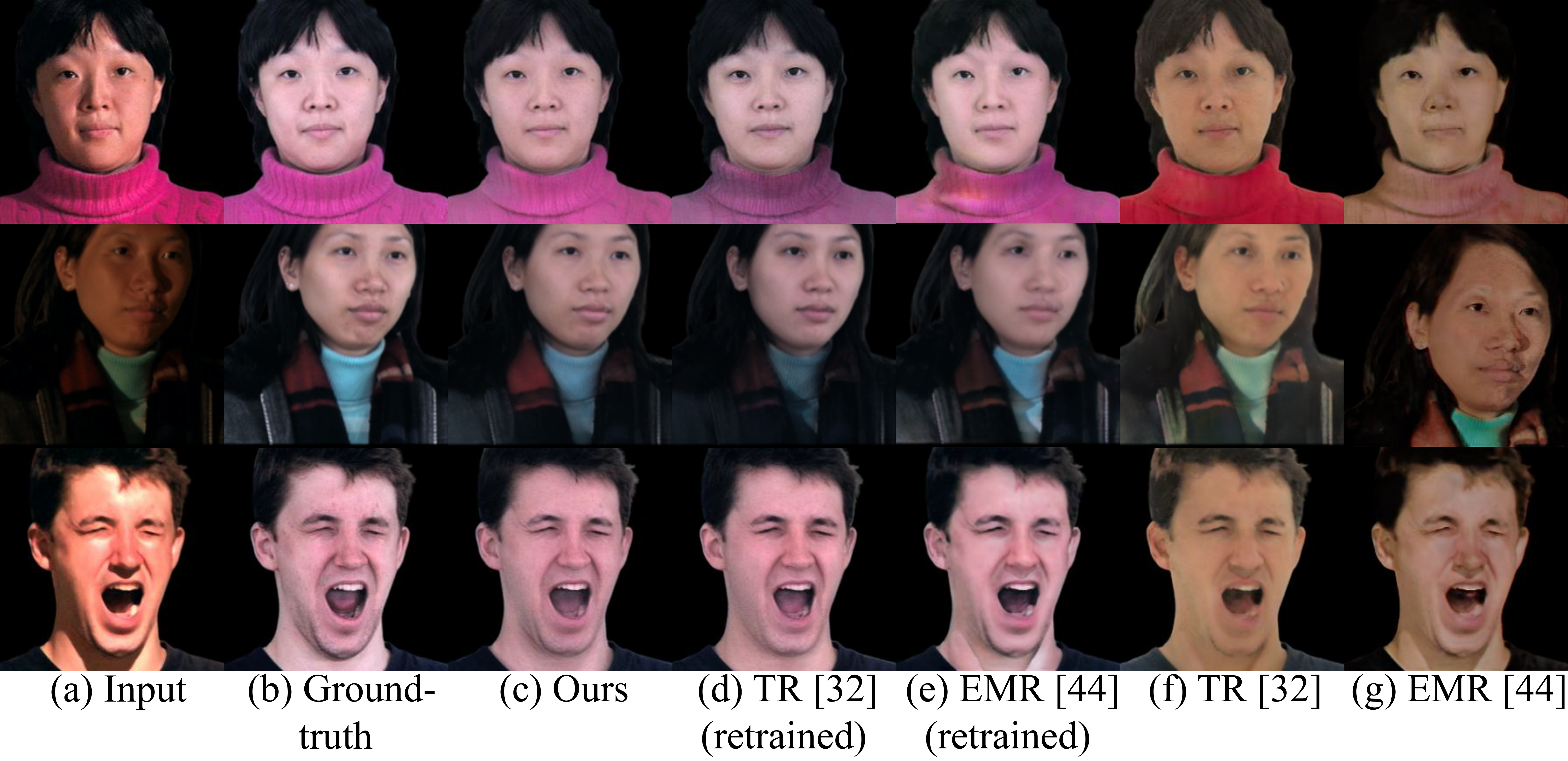}
    \caption{\textbf{Evaluation on our testing dataset.} Notice how our method is the only one to remove the scarf shadow in the middle row. A different crop of this subject was used for EMR since their method was trained on only face regions.}
    \label{testing_evaluation}
\end{figure}

\newcommand{\imgS}{0.14}

\begin{figure}
    \centering
    \includegraphics[width=\linewidth]{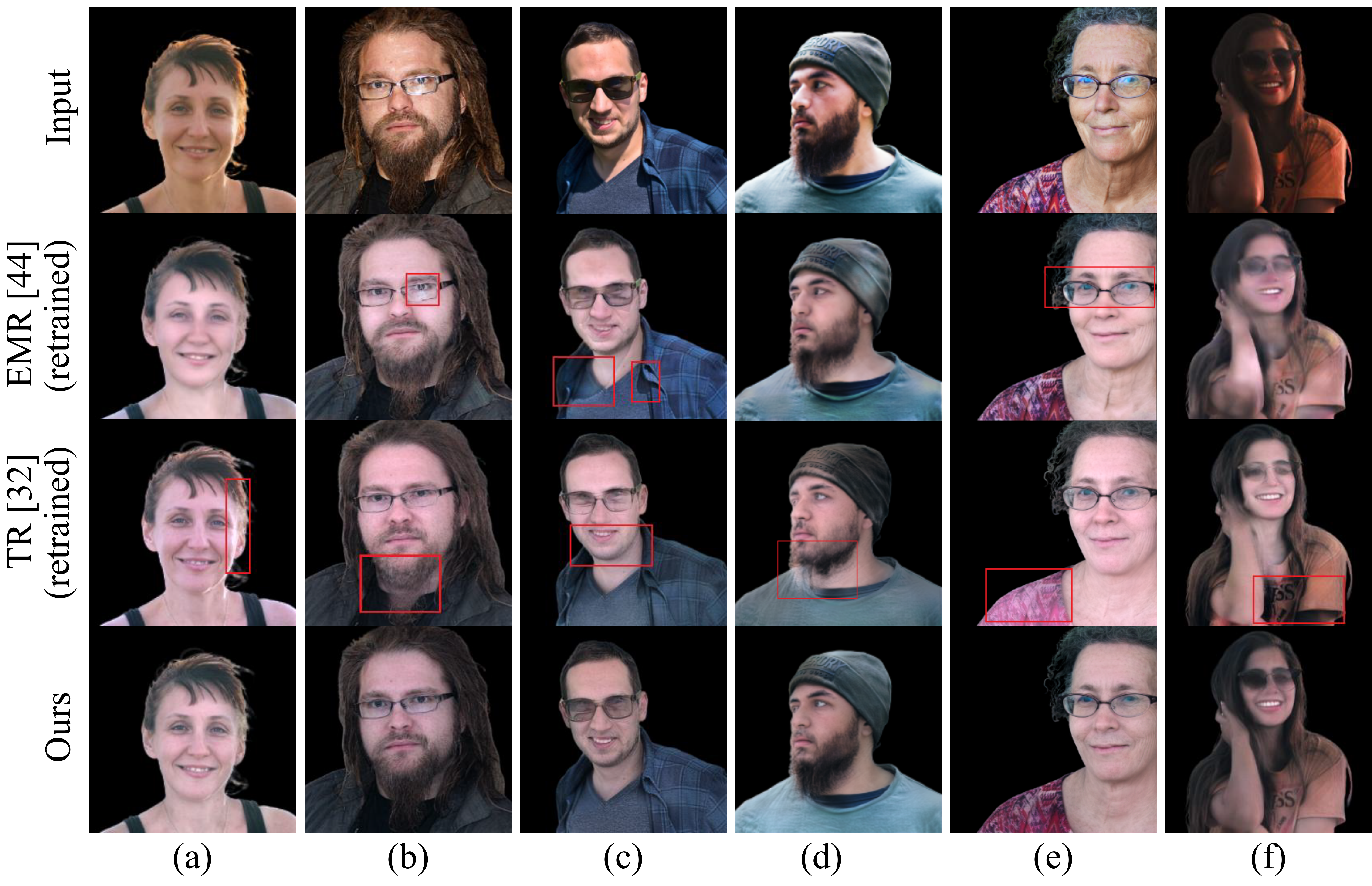}
    \caption{\textbf{Evaluations on in-the-wild images.} Red boxes emphasize areas in previous works improved by our method. Notice how our method provides the most consistent results in terms of delighting and texture preservation.}
    \label{in_the_wild}
\end{figure}

% \noindent{\textbf{Setups}}

\subsection{Implementation and Data Setup} \label{training_scheme}
We implement our model in PyTorch~\cite{paszke2019pytorch}, and train for 4 epochs with a learning rate of 0.0002 using the Adam optimizer~\cite{kingma2014adam}. All images are resized to $256\times256$ resolution, with pixel values normalized to $[-1, 1]$. The average running time of our network for delighting (without shading-offset prediction) is 25ms on a NVIDIA GTX 1080 GPU. 

To prepare the training data in our experiments, we apply random flips along the vertical axis, and random cropping with window sizes chosen uniformly from $[280, 480]$, where $480$ covers the entire image. When generating soft-shadow images in Eq.~\ref{guided_filter}, the large filter radius $\kappa$ is chosen randomly from $[7, 35]$ to increase robustness to different penumbra sizes, while the nose and mouth radius $\epsilon$ is fixed at 7.

% EMR (pretrained) uses a dataset of scanned head-to-shoulder subjects rendered with constant specular and roughness values, while TR (pretrained) uses an extensive dataset of full-body subjects captured in a dome of 331 LED-lights. Both EMR (retrained) and TR (retrained) use the same dataset and training scheme as our method.

% \subsection {Comparisons with the State-of-the-Art}
\noindent{\textbf{Prior Work.}}
We compare our method with the albedo prediction modules of two state-of-the-art methods: Explicit Multiple Reflectance
Channel Modeling (EMR)~\cite{wang2020single} and Total Relighting (TR)~\cite{pandey2021total}, which have demonstrated superior performance over previous related works. TR apply VGG-perceptual loss (see Eq.~\ref{l_perc}) on the estimated albedo, along with a Least-Squares GAN~\cite{mao2017least} discriminator to remove high-frequency shading. EMR use only $\ell_1$ loss on the albedo, and estimate the source illumination and a face parsing map as auxiliary tasks to improve training stability.

% EMR uses a dataset of scanned head-to-shoulder subjects rendered with constant specular and roughness values, while TR uses an extensive dataset of full-body subjects captured in a dome of 331 LED-lights. 

% Results from their pretrained models were generously provided to us by the authors using our evaluation dataset. But for a fair comparison, we trained implementations of their models on our dataset. 

% \noindent \textbf{Retrained models.}
% Besides the pretrained models from prior work, we trained implementations of EMR and TR on our dataset. For TR (retrained) and EMR (retrained), we implemented the albedo prediction network and loss functions as described in their papers~\cite{pandey2021total}~\cite{wang2020single}. For EMR (retrained), we used a similar network architecture as shown in Fig.~\ref{fig:architecture} with an added face-parsing layer, and with only $\ell_1$ loss applied to $\mathbf{I_{dlt}}$ and $\mathbf{I_{off}}$. We also kept our shading-offset decoder $\mathcal{D}_2$ for EMR (retrained) since our framework doesn't utilize environment maps. This has a similar effect to EMR's source lighting estimator. 

\noindent \textbf{Retrained models.} 
Besides the pretrained models from prior work, we \emph{retrained} EMR and TR on our dataset, denoted EMR (retrained) and TR (retrained) respectively. We based our implementations of their models on the albedo prediction networks and loss functions described in their papers~\cite{pandey2021total, wang2020single}. For EMR (retrained), We use our segmentations in Fig.~\ref{preprocessing} (e) as the ground-truth for face parsing. and since our pipeline doesn't utilize $360^\textrm{o}$ environment maps, we use our shading-offset decoder $\mathcal{D}_2$ as a substitute for their source lighting estimator. This allows us to use the lighting information found in our shading-offset images to perform this task, just as we do in our method (see Fig.~\ref{fig:architecture}).  

% \subsubsection{Qualitative Evaluations}
\subsection{Qualitative Evaluations}

In Fig.~\ref{testing_evaluation}, we compare our results qualitatively against these two methods on our testing dataset. From the results, we can see that our method outperforms both EMR and EMR (retrained) in terms of shadow and specular removal (middle and bottom rows). Our  method  also  recovers large-scale textures, whereas TR and TR (retrained) are prone to incorrect colour estimations as can be seen with the magenta shirt in the top row. Although TR handles most shading effects on the face, it still leaves shadow boarders in some extreme cases (bottom row, (f)), while our proposed method can remove them effectively, indicating that our method is generalized to harsh cases.

% We show results on in-the-wild images in Fig.~\ref{in_the_wild}. Here, we see that our method is able to delight subjects under a wide range of complex illuminations, while also preserving important details such as clothing patterns, facial hair, and hair covering the face. Our method can also handle variant conditions more gracefully than other works. For example, TR (retrained) often removes important content details such as the beard in (d), while EMR (retrained) preserves shadow boarders and sharp reflections (second row, (d) and (e)), and is unstable when images contain large non-face components (second row, (f)). In (c), we demonstrate another failure case of TR (retrained) where light is coming from behind the subject. Although this lighting condition is missing from our dataset, our model generalizes to this case very well by the contribution of  our soft-shadow loss (see Sec.~\ref{ablation_section} for further insight).  

We show results on in-the-wild images in Fig.~\ref{in_the_wild}. Here, we see that our method is able to delight subjects under a wide range of complex illuminations, while also preserving important details such as clothing patterns and facial hair. Our method handles variant conditions more gracefully than other works. For example, TR (retrained) often removes important content details such as beards ((b) and (d)), while EMR (retrained) preserves shadow boarders and sharp reflections ((b) and (c)), and is unstable when images contain large non-face components (f). In (a), we demonstrate another failure case of TR (retrained) where light is coming from behind the subject. Although this lighting condition is missing from our dataset, our model generalizes to this case very well by the contribution of our soft-shadow loss (see Sec.~\ref{ablation_section} for further insight).  

% \subsubsection {Quantitative Evaluations}
\subsection {Quantitative Evaluations}
\setlength{\tabcolsep}{6pt}
\begin{center}
\begin{table}[htb]
\begin{center}
\caption{Results of delighting on our testing dataset. Arrows indicate whether loss is minimized ($\downarrow$) or maximized ($\uparrow$). Our method outperforms prior works on all metrics.}
\label{qunantitative}
% \resizebox{\columnwidth}{!}{
\begin{tabular}{ p{3cm}|cccc}
 \hline
 & \multicolumn{4}{c}{\textbf{Metric}} \\
 \hline
 \textbf{Method} & \textbf{RMSE$\downarrow$} & \textbf{SSIM$\uparrow$} & \textbf{li-SSIM$\uparrow$} & \textbf{LPIPS$\downarrow$}\\
 \hline
 EMR (retrained) & 0.047 & 0.940 & 0.949 & 0.047\\
 \hline
 TR (retrained) & 0.056 & 0.934 & 0.948 & 0.048\\ 
 \hline
 Proposed & \textbf{0.044} & \textbf{0.946} & \textbf{0.955} & \textbf{0.037}\\
 \hline
\end{tabular}
% }

\end{center}
\end{table}
\end{center}

Quantitative evaluations on our testing dataset are shown Tab.~\ref{qunantitative}, where we compare our delighting performance with TR (retrained) and EMR (retrained) using the following metrics: root mean squared error (RMSE), structural similarity (SSIM)~\cite{wang2004image} luminance-invariant SSIM (li-SSIM) and learned perceptual image patch similarity (LPIPS)~\cite{zhang2018unreasonable} (version 0.1). Our li-SSIM is like traditional SSIM with luminance parameter $\alpha = 0$. This is meant to avoid biases to large-scale colour variations, penalizing contrast and structural errors only. From the results we can see that our method outperforms other works across all metrics. Notably, TR gains a significant improvement on the li-SSIM metric over SSIM, but is still behind our method. These results reflect consistent improvements in most of our qualitative images. More of our results can be found in the supplementary document.

\subsection {Ablation study} \label{ablation_section}

Our baseline model consists of our network trained using only the basic loss $\mathcal{L_{\mathbf{dlt}}}$ (see Eq.~\ref{dltEq}). We add each of our proposed losses onto this model to evaluate their contributions. Quantitative results of all our ablations using our testing dataset are shown in Tab.~\ref{ablations_quantitative}.

\begin{figure}
\centering
\includegraphics[width=1.0\textwidth]{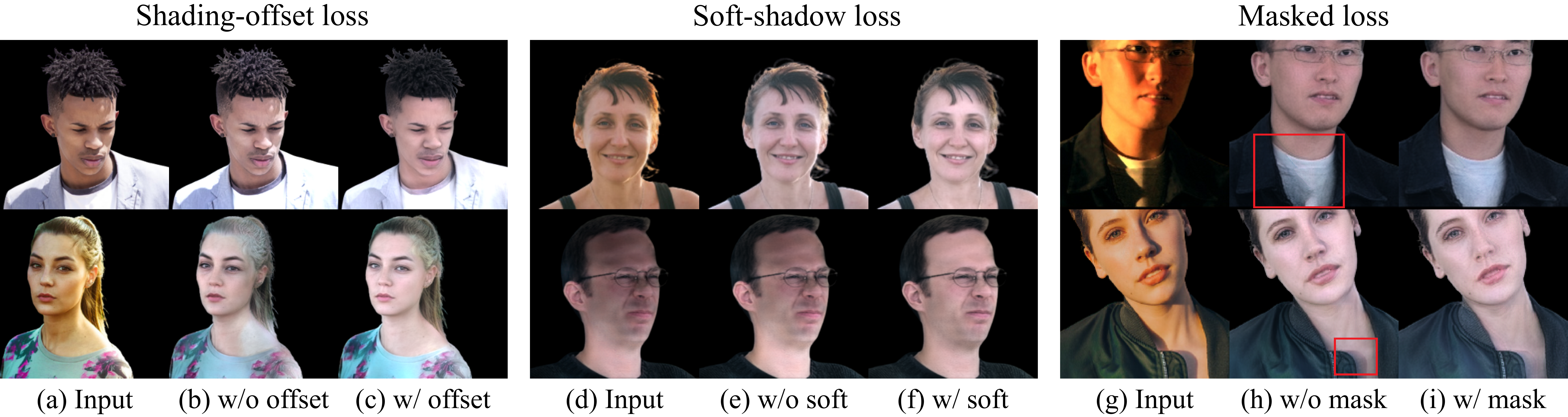}
\caption{\textbf{Ablation study} on (a, b, c) shading-offset loss, (d, e, f) soft-shadow loss and (g, h, i) masked loss. Red boxes in (h) highlight shadows.}
\label{ablations}
\end{figure}

\noindent \textbf{Shading-offset loss.} We show the benefits provided by our offset decoder $\mathcal{D}_2$ by training an additional model with shading-offset loss $\mathcal{L_{\mathbf{off}}}$ (see Eq.~\ref{offEq}) added to our baseline. Qualitative results in Fig.~\ref{ablations} (c) demonstrate more stable texture recovery and light removal when using offset loss. The likely reason is that it could make the learned latent features more discriminative between different colour variations caused by shading and texture. In Tab~\ref{ablations_quantitative} (\textbf{Testing Dataset} block), we see that our model with offset loss performs better than our baseline across all metrics, especially RMSE and LPIPS, indicating improved large-scale texture recovery. 

% \begin{figure*}
% \centering
% \includegraphics[width=1.0\textwidth, trim={1.5cm 23.3cm 1.5cm 0.8cm},clip]{images/newImages/combine_image_6.pdf}
% \caption{\textbf{Ablation study} on (a, b, c) shading-offset loss, (d, e, f) soft-shadow loss and (g, h, i) masked loss. Red boxes in (h) highlight shadows.}
% \label{ablations}
% \end{figure*}

\noindent \textbf{Soft-shadow loss.} \label{soft-shadow-ablation} We add the soft-shadow loss $\mathcal{L_{\mathbf{soft\mbox{-}dlt}}} + \mathcal{L_{\mathbf{soft\mbox{-}off}}}$ (see Eq.~\ref{soft-shading-loss_alb} \& ~\ref{soft-shading-loss_off}) to our shading-offset ablation. While our perceptual loss in  $\mathcal{L}_{\mathbf{dlt}}$ allows for stable delighting under harsh illumination, it can lead to over-fitting due to the mostly front-facing directional lights from our training data, and can't cope with diffuse illuminations very well. From the results in Fig.~\ref{ablations} (f), we can see that adding our soft-shadow regularization not only improves performance on diffuse illumination conditions (bottom row), but also enables our model to recognize rare edge cases such as in the top row, where we observe that strong light diffractions through the hair and around the left cheek are removed.

To better quantify the benefits of our soft-shadow regularization, we create a new testing dataset (dubbed as \textbf{Alt. Lighting}) using the face relighting method of Hou \etal~\cite{hou2021towards} to obtain sufficiently variant testing samples. The de-lit images from our original testing dataset were relit under 13 lighting conditions from a wide range of angles not seen in our training dataset. Quantitative results in Tab.~\ref{ablations_quantitative} (\textbf{Alt. Lighting} block) demonstrates how our soft-shadow loss improves robustness to these conditions, 
particularly on the RMSE and LPIPS metrics. 

\noindent \textbf{Masked loss.} We add our masked loss $\mathcal{L_{\mathbf{msk}}}$ (Eq. \ref{hf_mask}) to our soft-shadow ablation. As the results in Fig. \ref{ablations} (i) show, our masked loss improves the removal of complex shadows that are intertwined with clothing features. It is noteworthy that although the model without $\mathcal{L_{\mathbf{msk}}}$ can handle a wide-range of harsh illumination conditions situated around the face area, abnormal cast shadows, particularly around the torso are often preserved. By adding extra importance to these regions via $\mathcal{L_{\mathbf{msk}}}$, our model reduces the rate at which they get preserved in the result. Quantitative results in Tab.~\ref{ablations_quantitative} show that our model without $\mathcal{L_{\mathbf{msk}}}$ outperforms our proposed method on most metrics, although we can see the benefit of $\mathcal{L_{\mathbf{msk}}}$ in our testing cases visually. This could mean the weighted mask creates a bias towards these relatively small regions in the loss function, at the expense of larger image structures.

\begin{center}
\begin{table}[htb]
\begin{center}
\caption{Ablation results on our loss functions. $\mathcal{L_{\mathbf{soft}}} = \mathcal{L_{\mathbf{soft\mbox{-}dlt}}} + \mathcal{L_{\mathbf{soft\mbox{-}off}}}$. We evaluate against two datasets: Testing Dataset (left block) and Alt. Lighting (right block), where our testing dataset is the same one used in Tab.~\ref{qunantitative}. }
\label{ablations_quantitative}
\begin{tabular}{ cc|ccc|ccc}
 \hline
 \multicolumn{2}{c|}{} & \multicolumn{6}{c}{\textbf{Metric}} \\
 \hline
 \multicolumn{2}{c|}{}& \multicolumn{3}{c|}{\textbf{Testing Dataset}} & \multicolumn{3}{c}{\textbf{Alt. Lighting}}\\
 \hline
 \multicolumn{2}{c|}{\textbf{Method}} & \textbf{RMSE$\downarrow$} & \textbf{SSIM$\uparrow$} & \textbf{LPIPS$\downarrow$} & \textbf{RMSE$\downarrow$} &
 \textbf{SSIM$\uparrow$} &
 \textbf{LPIPS$\downarrow$}\\
 \hline
 (A)&$\mathcal{L_{\mathbf{dlt}}}$ & 0.056 & 0.943 & 0.042 & 0.040 & 0.979 & 0.052\\
 \hline
 (B)&(A)$+\mathcal{L_{\mathbf{off}}}$ & 0.046 & \textbf{0.949} & \textbf{0.035} & 0.041 & 0.979 & 0.052 \\ 
 \hline
 (C)&(B)$+\mathcal{L_{\mathbf{soft}}}$ & 0.046 & \textbf{0.949} & \textbf{0.035} & \textbf{0.026} & \textbf{0.985} & \textbf{0.033}\\
 \hline
 (D)&(C)$+\mathcal{L_{\mathbf{msk}}}$ & \textbf{0.044} & 0.946 & 0.037 & \textbf{0.026} & 0.984 & 0.037\\
 \hline
\end{tabular}
\end{center}
\end{table}
\end{center}

\begin{figure}
    \centering
    \includegraphics[width=0.88\linewidth]{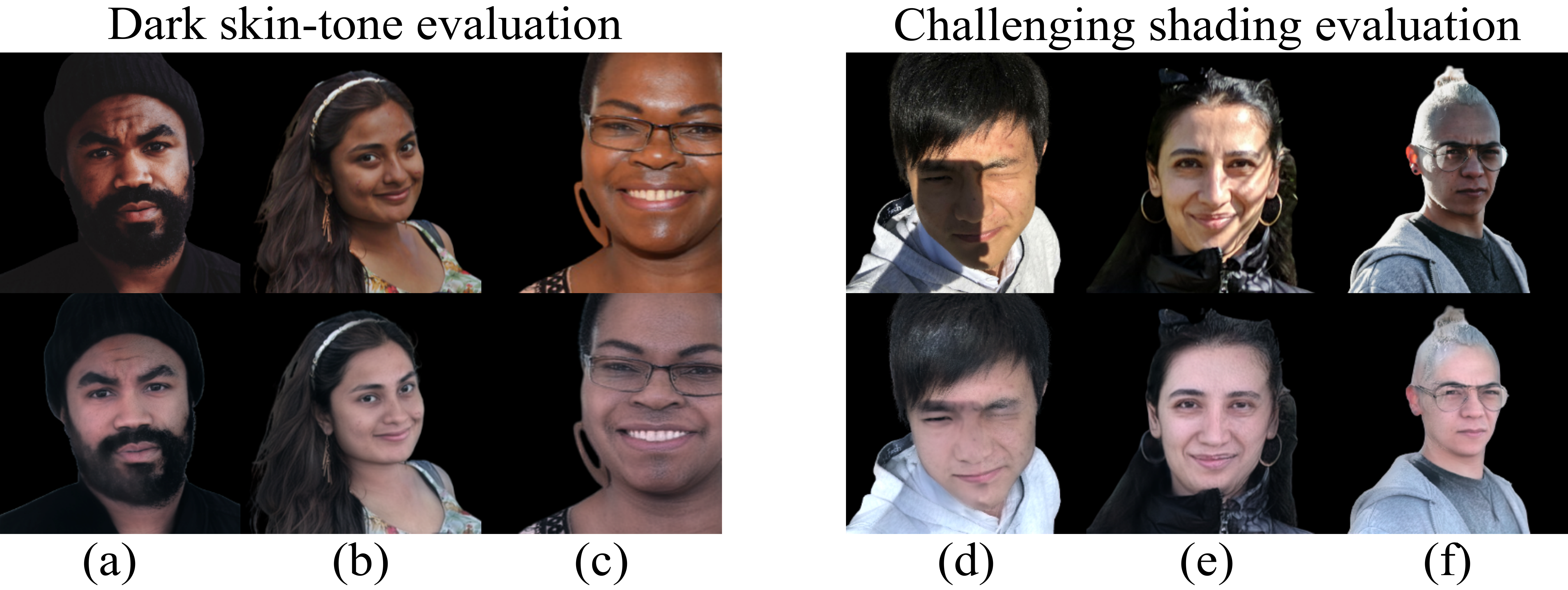}
    \caption{\textbf{Edge cases:} Our results on images under-represented in our training data. Source images of (d) \& (e) are from the evaluation dataset of \cite{zhang2020portrait}.}
    \label{fig_edge_cases}
\end{figure}

\subsection{Edge Cases} \label{edge_cases}

While the Multi-PIE \cite{gross2010multi} dataset exhibits much diversity in terms of clothing, expression and identity, it nonetheless has significant domain biases such as a 95\% European/Asian demographic captured under mostly front-facing directional lightings. This can lead to failure cases when presented with dark skin albedos, and challenging illuminations such as light coming from extreme angles or shadows cast by foreign objects. In this section, we offer a qualitative evaluation on these under-represented cases.

We present our evaluation on people with dark skin in Fig. \ref{fig_edge_cases} (a-c). From the results, we can see that our method is capable of recovering dark skin textures, although noticeable lightening is observed.

In Fig \ref{fig_edge_cases} (d-f), we demonstrate our results on (d, e) irregularly shaped shadows, and (f) lighting from predominantly behind the subject. While no such images were present in our training data, our model nonetheless generalizes to these cases surprisingly well with a few minor artifacts.

%% file: applications.tex
\begin{figure}
    \centering
    \includegraphics[width=0.65\linewidth]{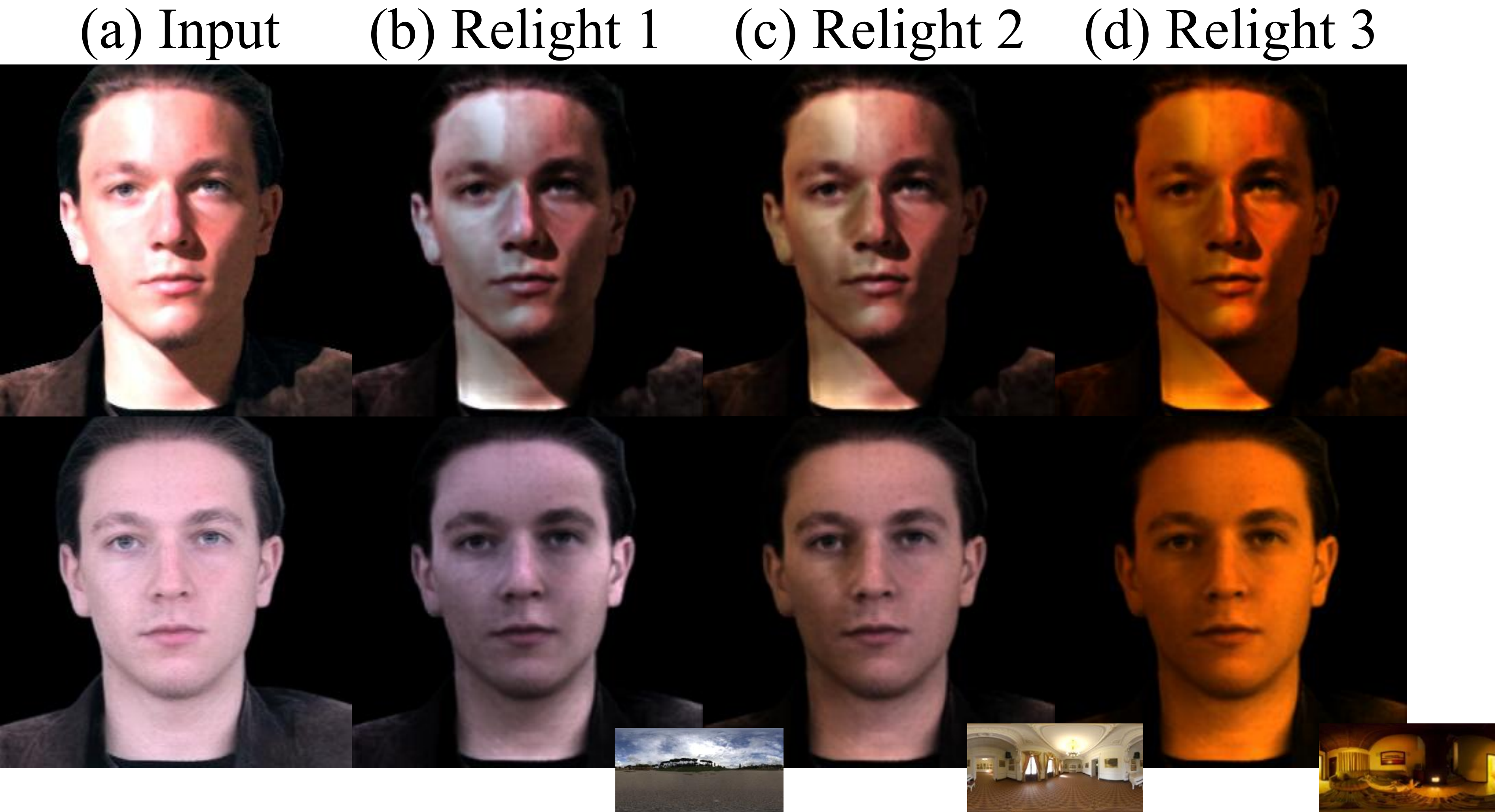}
    \caption{\textbf{Relighting.} The results of face relighting~\cite{hou2021towards} (top row) before and (bottom row) after our delighting step.}
    \label{relighting}
\end{figure}

\begin{figure}
    \centering
    \includegraphics[width=0.6\textwidth]{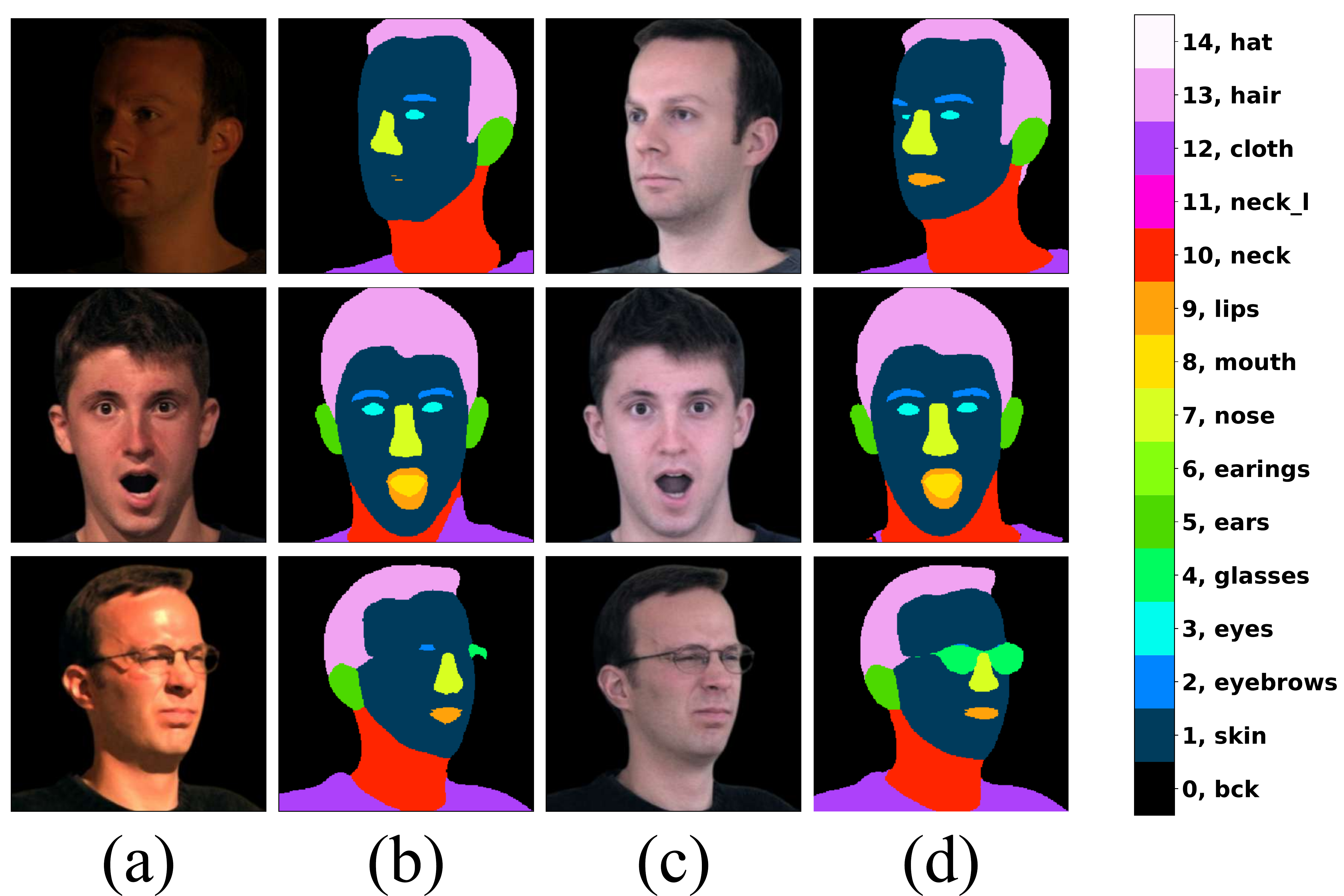}
    \caption{\textbf{Face Parsing:} Here we show (a) the input image with (b) its semantic parsing output, and (c) our de-lit image with (d) its semantic parsing output. }
    \label{face-parsing}
\end{figure}

% \noindent \textbf{Face Relighting.} Fig.~\ref{relighting} illustrates the benefits of our delighting as a preprocessing step for other relighting methods. Through our shadow removal and colour normalization, we achieve more accurate results using the relighting method of Hou \etal~\cite{hou2021towards}. More relighting examples can be found in our supplementary document and video. 

\noindent \textbf{Face Relighting.} Fig.~\ref{relighting} illustrates the benefits of our delighting as a preprocessing step for other relighting methods. Through our shadow removal and colour normalization, we achieve more natural looking results on HDRI environments. Relighting is performed using the method of Hou \etal \cite{hou2021towards}, which we adapt for environment map renderings by taking a weighted sum of 512 unique illuminations (similar to \cite{debevec2000acquiring}). 

\noindent \textbf{Face Parsing.} Fig.~\ref{face-parsing} illustrates the importance of delighting when performing semantic segmentation on portraits. Our model offers significant performance gains to this task by enhancing the visual clarity of glasses, eyes and mouth regions. Also noteworthy is the neck shadow (middle-row, (a)), where without delighting, the neck region outside is classified as clothing (middle-row, (b)). The face parser used is a BiSeNet model~\cite{yu2018bisenet} trained on the CelebAMask-HQ dataset~\cite{CelebAMask-HQ}.   

%% file: Conclusion.tex
We propose a deep neural network for delighting portrait images under a wide range of illumination conditions. Texture recovery and generalization is improved via estimating shading-offset images, using soft-shadow variants of the input, and our weighted loss function. Each contribution greatly improves delighting performance over the previous work in terms of removing shading features and preserving image content. We tested our model as a useful preprocessing tool for other computer vision tasks.

Failure cases can arise due to biases in the training data as outlined in Sec. \ref{edge_cases}, so future work must consider fairness in terms of physical traits and lighting variations when creating and using datasets. To prevent softening of dark textures (Fig. \ref{fig_edge_cases} (f)), future work can also focus on estimating global image statistics, such as ambient light intensity, which will place a lower bound on the darkest regions of the image caused by shading.\newline

\noindent \textbf{Acknowledgments} This work was supported by the Entrepreneurial University Programme from the Tertiary Education Commission, and MBIE Smart Idea Programme by Ministry of Business, Innovation and Employment in New Zealand. We thank all image providers, including Flickr users: "Debarshi Ray", "5of7" and "photographer695", whose photographs were cropped, and processed by our neural network. Sources are provided in the supplementary material.  

% We also thank Chloe Legendre and Zhibo Wang for running their trained models on our dataset. 

%% file: supplementary.tex
\section{OLAT and Target synthesis}
In sec. 3.1 of our paper, we describe how we process our data for training. Here, we provide a deeper overview of how we remove the effect of the room lights in the Multi-PIE \cite{gross2010multi}  flash images, and synthesize the target de-lit image.

Since the images with no camera-flashes (room-lit) are available, we can simply remove their effect with the following equation. 

\begin{equation}
    \mathbf{I_{ratio} = 1 - (\frac{L(I_{room lights})}{L(I_{flash})})}
\end{equation}
\begin{equation}
    \mathbf{I_{no\_ambient} = I_{ratio} \odot I_{flash}} 
\end{equation}

\noindent Where $\odot$ is pixel-wise multiplication, $\mathbf{L(A)}$ refers to the luminance channel of $\mathbf{A}$ in Lab color space. $\mathbf{I_{roomlights}}$ and $\mathbf{I_{flash}}$ are the room-lit and flash images respectively.

Another problem is that the room lights create sharp specular reflections on the faces of some subjects. Since the brightness of these regions is near full (images were taken with a LDR camera), removing these from our flash images results in dark blemishes across the face (Fig. \ref{OLAT_synth} (c)). We detect these regions using the following equation:

\begin{equation} \label{detectSpecular}
    \mathbf{I_{specular} = min(1, \frac{{I_{roomlights}}^2}{I_{average}})^4)}, 
\end{equation}

\noindent where $\mathbf{I_{average}}$ is the mean average over all flash images. We correct these small regions using Navier-Strokes \cite{bertalmio2001navier} inpainting to produce $\mathbf{I_{OLAT}}$. An illustration of each step is shown in Fig. \ref{OLAT_synth}.

To synthesize the target de-lit image, we approximate ambient lighting as the average over all our 18 OLAT images, producing $\mathbf{I_{dlt}}$. From here, we modify $\mathbf{I_{dlt}}$ using the following equation:

\begin{equation} \label{delitImage}
    \mathbf{L(I_{dlt})} = \mathbf{L(I_{dlt})} + 6\mathbf{L(I_{room\_nospec})}, 
\end{equation}

\noindent where $\mathbf{I_{room\_nospec}}$ is $\mathbf{I_{roomlights}}$ with specularities removed (like $\mathbf{I_{OLAT}}$). Adding the luminance of the room-lit image back into $\mathbf{I_{dlt}}$ makes the lighting appear more uniform.

\newcommand{\imgOLAT}{0.24}
\begin{figure}
% \captionsetup[subfigure]{format=empty}
\centering
% \captionsetup[subfigure]{labelformat=empty}
\begin{subfigure}[t]{\imgOLAT\linewidth}
    \includegraphics[width=\textwidth]{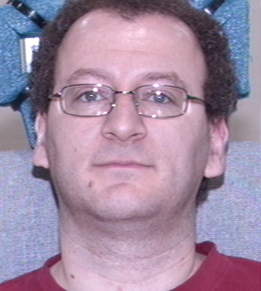}
    \caption{$\mathbf{I_{flash}}$}
\end{subfigure}
\begin{subfigure}[t]{\imgOLAT\linewidth}
    \includegraphics[width=\textwidth]{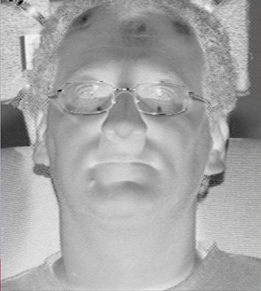}
    \caption{$\mathbf{I_{ratio}}$}
\end{subfigure}
\begin{subfigure}[t]{\imgOLAT\linewidth}
    \includegraphics[width=\textwidth]{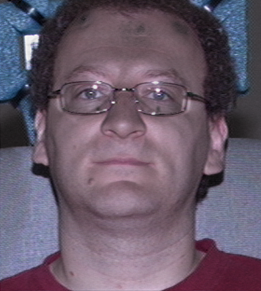}
    \caption{$\mathbf{I_{no\_ambient}}$}
\end{subfigure}
\begin{subfigure}[t]{\imgOLAT\linewidth}
    \includegraphics[width=\textwidth]{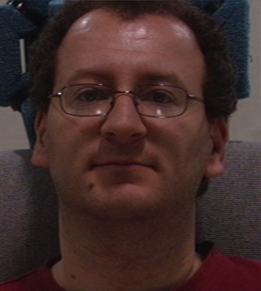}
    \caption{$\mathbf{I_{roomlights}}$}
\end{subfigure}
\begin{subfigure}[t]{\imgOLAT\linewidth}
    \includegraphics[width=\textwidth]{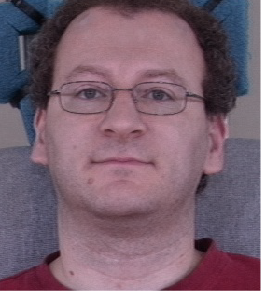}
    \caption{$\mathbf{I_{dlt}}$}
\end{subfigure}
\begin{subfigure}[t]{\imgOLAT\linewidth}
    \includegraphics[width=\textwidth]{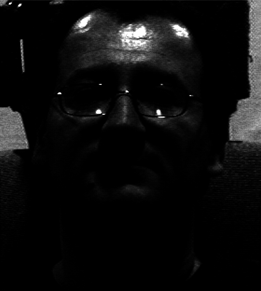}
    \caption{$\mathbf{I_{specular}}$}
\end{subfigure}
\begin{subfigure}[t]{\imgOLAT\linewidth}
    \includegraphics[width=\textwidth]{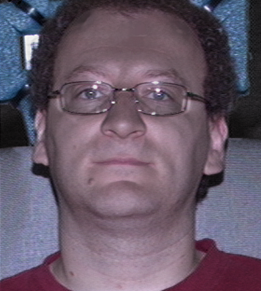}
    \caption{$\mathbf{I_{OLAT}}$}
\end{subfigure}
\begin{subfigure}[t]{\imgOLAT\linewidth}
    \includegraphics[width=\textwidth]{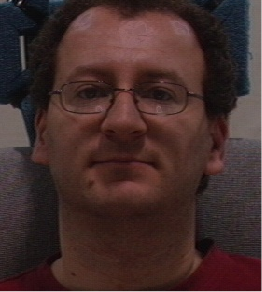}
    \caption{$\mathbf{I_{room{\_}nospec}}$}
\end{subfigure}
\caption{Illustration of our data synthesis stage.}
\label{OLAT_synth}
\end{figure}

\section{More comparisons}

In sec. 4.2 of our paper, we showed qualitative comparisons against TR and EMR against our testing dataset. However, due to different ground-truth modalities (\ie albedo quality and targeted regions), we don't compare quantitatively against the pretrained models of TR and EMR in sec. 4.3. As an extension to Fig. 4 in the main paper, we provide a more in-depth qualitative comparison.

\noindent \textbf{Upper-body portraits} In Fig. \ref{half-body-comparisons} We provide examples from our testing dataset of images containing significantly large clothing regions. Since EMR wasn't trained on such examples, we do not present their results in this section, and instead show only EMR (retrained).

\noindent \textbf{Face portraits} In Fig. \ref{face-crop-comparison}, we show more comparisons of images cropped around the face region. Here, results of all prior works mentioned in the paper are shown.

% \section{More masked-loss examples} 
% In Fig. \ref{masked_ablation}, we show more qualitative examples of our masked loss function (see sec. 3.2 of our paper). Notice how our model trained with masked-loss can remove or soften challenging and irregular shading effects. The last row is an image from our testing dataset, while rows 1-5 are in-the-wild images with no ground-truth. 

\section{More soft-shadow loss examples}
In Fig. \ref{soft-ablation}, we show more qualitative examples of our soft-shadow regularization  (see sec. 3.2 of our paper). Input images are from our \textbf{Alt. Lighting} dataset (see sec. 4.4 of our paper).

\section{Shadow Removal}
The paper of Zhang \etal \cite{zhang2020portrait} focuses only on a subset of the delighting task: shadow removal/softening, hence we did not compare against their method in the main paper. We include this comparison in Fig. \ref{shadow_removal_2} so that viewers can qualitatively assess our shadow removal. In their work, they design an extensive dataset of shadowed/un-shadowed image pairs, in which shadows are cast by irregularly shaped objects outside the image (foreign shadows). While no such shadows were present in our training data, our model holds up well, while also achieving the overall delighting task including reflection removal.

\newcommand{\imgHalf}{0.16}
\begin{figure*}
\centering
\begin{subfigure}[t]{\imgHalf\textwidth}
    \includegraphics[height=\textwidth,width=\textwidth]{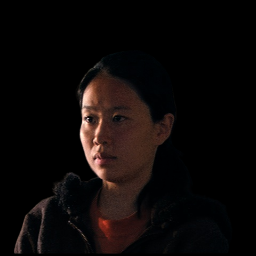}
    \includegraphics[height=\textwidth,width=\textwidth]{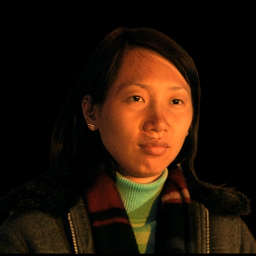}
    \includegraphics[height=\textwidth,width=\textwidth]{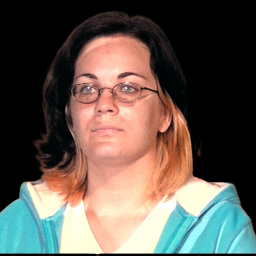}
    \includegraphics[height=\textwidth,width=\textwidth]{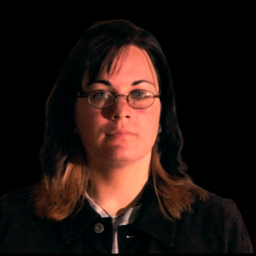}
    \includegraphics[height=\textwidth,width=\textwidth]{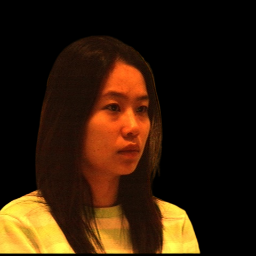}
    \includegraphics[height=\textwidth,width=\textwidth]{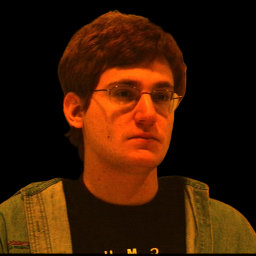}
    \caption{\centering Input}
\end{subfigure}
\begin{subfigure}[t]{\imgHalf\textwidth}
    \includegraphics[height=\textwidth,width=\textwidth]{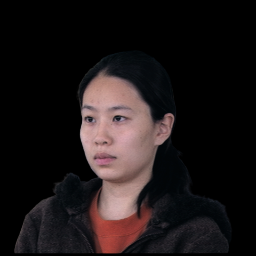}
    \includegraphics[height=\textwidth,width=\textwidth]{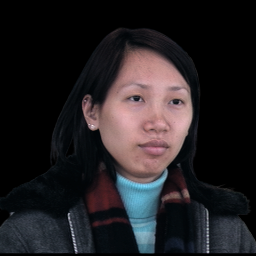}
    \includegraphics[height=\textwidth,width=\textwidth]{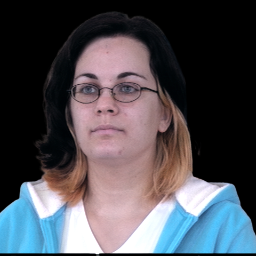}
    \includegraphics[height=\textwidth,width=\textwidth]{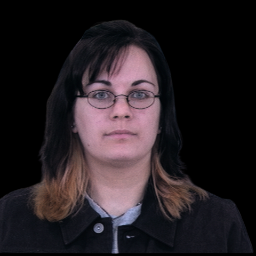}
    \includegraphics[height=\textwidth,width=\textwidth]{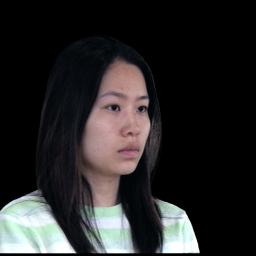}
    \includegraphics[height=\textwidth,width=\textwidth]{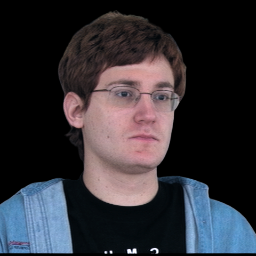}
    \caption{Ground truth}
\end{subfigure}
\begin{subfigure}[t]{\imgHalf\textwidth}
    \includegraphics[height=\textwidth,width=\textwidth]{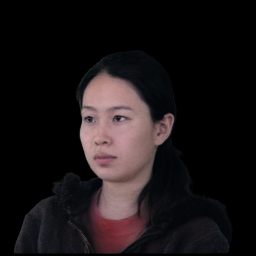}
    \includegraphics[height=\textwidth,width=\textwidth]{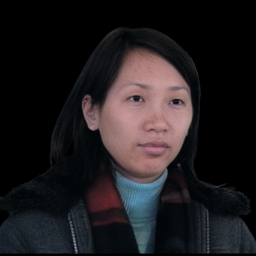}
    \includegraphics[height=\textwidth,width=\textwidth]{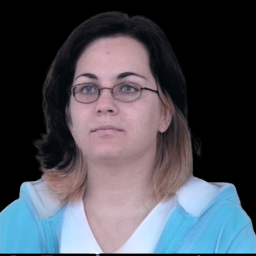}
    \includegraphics[height=\textwidth,width=\textwidth]{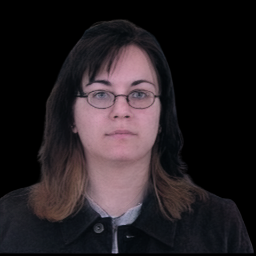}
    \includegraphics[height=\textwidth,width=\textwidth]{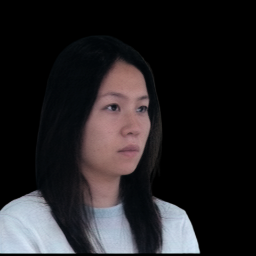}
    \includegraphics[height=\textwidth,width=\textwidth]{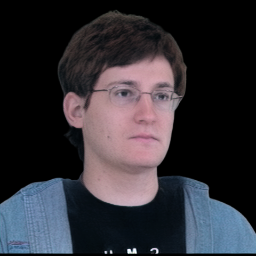}
    \caption{Ours}
\end{subfigure}
\begin{subfigure}[t]{\imgHalf\textwidth}
    \includegraphics[height=\textwidth,width=\textwidth]{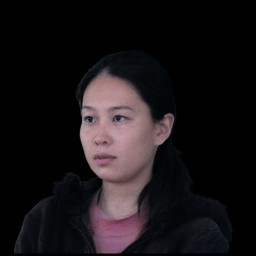}
    \includegraphics[height=\textwidth,width=\textwidth]{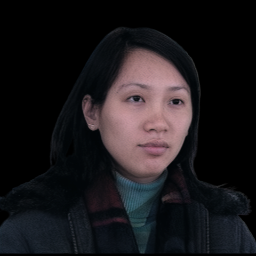}
    \includegraphics[height=\textwidth,width=\textwidth]{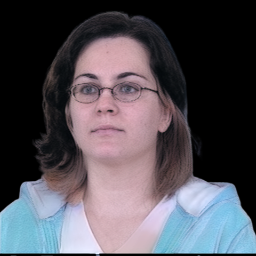}
    \includegraphics[height=\textwidth,width=\textwidth]{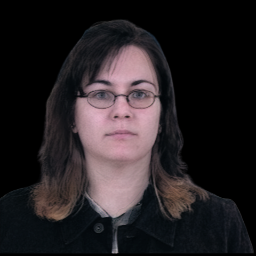}
    \includegraphics[height=\textwidth,width=\textwidth]{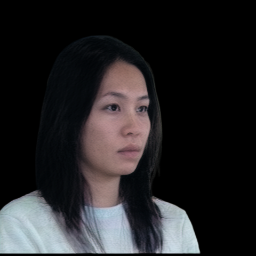}
    \includegraphics[height=\textwidth,width=\textwidth]{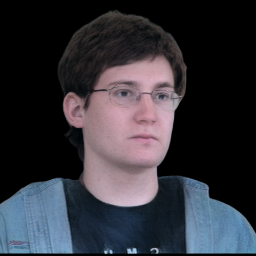}
    \caption{\centering TR (retrained)}
\end{subfigure}
\begin{subfigure}[t]{\imgHalf\textwidth}
    \includegraphics[height=\textwidth,width=\textwidth]{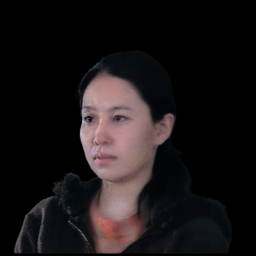}
    \includegraphics[height=\textwidth,width=\textwidth]{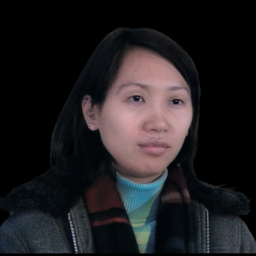}
    \includegraphics[height=\textwidth,width=\textwidth]{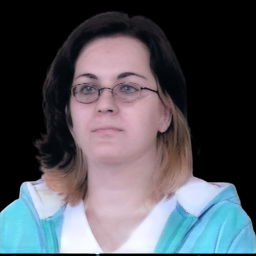}
    \includegraphics[height=\textwidth,width=\textwidth]{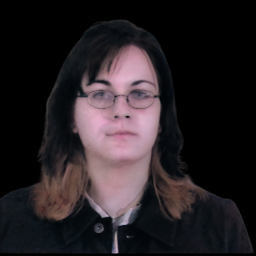}
    \includegraphics[height=\textwidth,width=\textwidth]{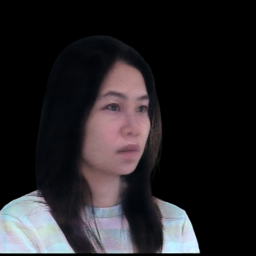}
    \includegraphics[height=\textwidth,width=\textwidth]{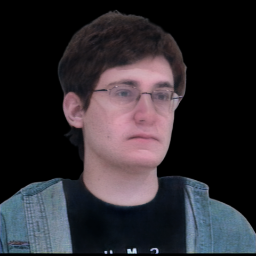}
    \caption{\centering EMR (retrained)}
\end{subfigure}
\begin{subfigure}[t]{\imgHalf\textwidth}
    \includegraphics[height=\textwidth,width=\textwidth]{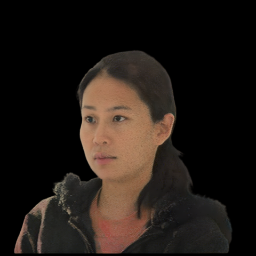}
    \includegraphics[height=\textwidth,width=\textwidth]{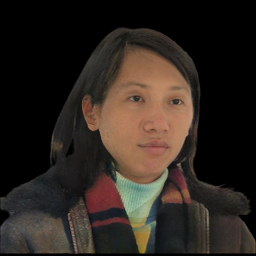}
    \includegraphics[height=\textwidth,width=\textwidth]{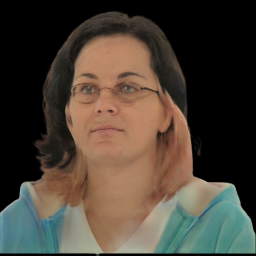}
    \includegraphics[height=\textwidth,width=\textwidth]{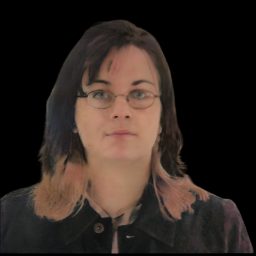}
    \includegraphics[height=\textwidth,width=\textwidth]{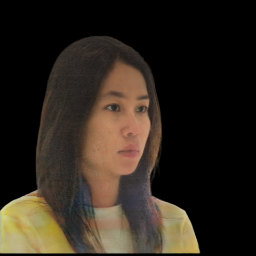}
    \includegraphics[height=\textwidth,width=\textwidth]{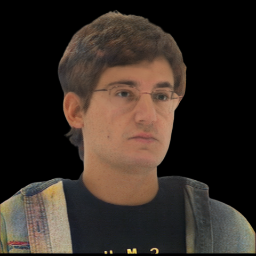}
    \caption{\centering TR}
\end{subfigure}
\caption{We show more comparisons against other methods using upper body images from our testing dataset.}
\label{half-body-comparisons}
\end{figure*}

\newcommand{\imgCrop}{0.13}
\begin{figure*}
\centering
\begin{subfigure}[t]{\imgCrop\textwidth}
    \includegraphics[height=\textwidth,width=\textwidth]{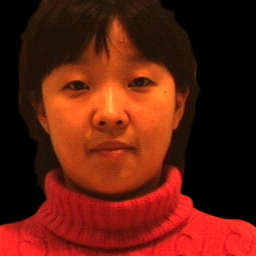}
    \includegraphics[height=\textwidth,width=\textwidth]{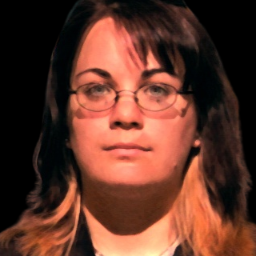}
    \includegraphics[height=\textwidth,width=\textwidth]{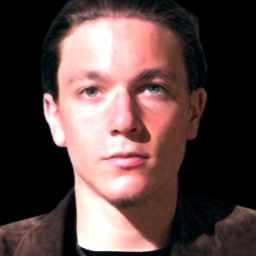}
    \includegraphics[height=\textwidth,width=\textwidth]{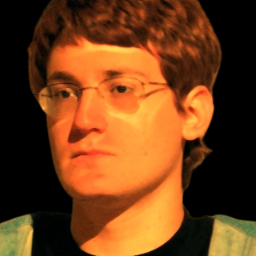}
    \includegraphics[height=\textwidth,width=\textwidth]{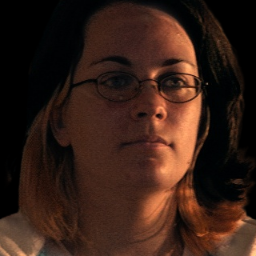}
    \includegraphics[height=\textwidth,width=\textwidth]{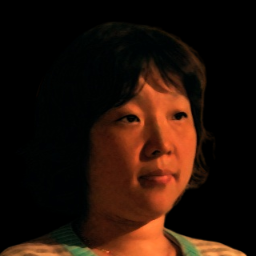}
    \caption{\centering Input}
\end{subfigure}
\begin{subfigure}[t]{\imgCrop\textwidth}
    \includegraphics[height=\textwidth,width=\textwidth]{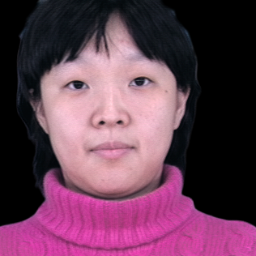}
    \includegraphics[height=\textwidth,width=\textwidth]{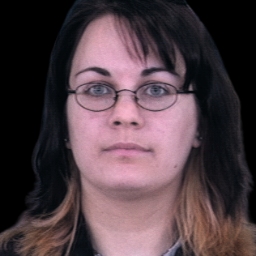}
    \includegraphics[height=\textwidth,width=\textwidth]{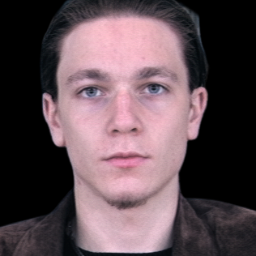}
    \includegraphics[height=\textwidth,width=\textwidth]{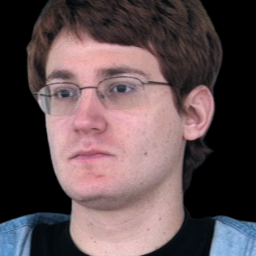}
    \includegraphics[height=\textwidth,width=\textwidth]{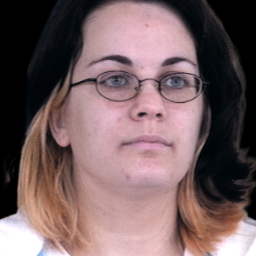}
    \includegraphics[height=\textwidth,width=\textwidth]{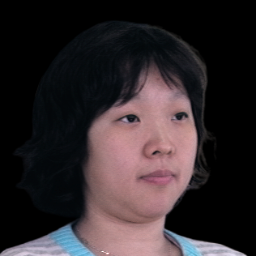}
    \caption{Ground truth}
\end{subfigure}
\begin{subfigure}[t]{\imgCrop\textwidth}
    \includegraphics[height=\textwidth,width=\textwidth]{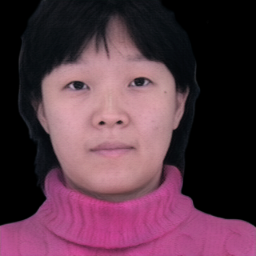}
    \includegraphics[height=\textwidth,width=\textwidth]{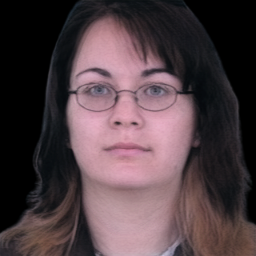}
    \includegraphics[height=\textwidth,width=\textwidth]{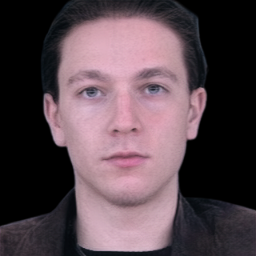}
    \includegraphics[height=\textwidth,width=\textwidth]{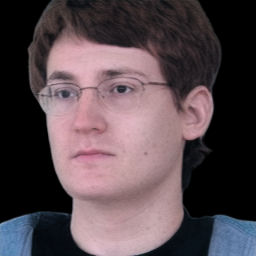}
    \includegraphics[height=\textwidth,width=\textwidth]{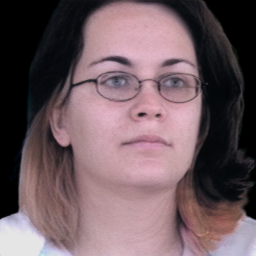}
    \includegraphics[height=\textwidth,width=\textwidth]{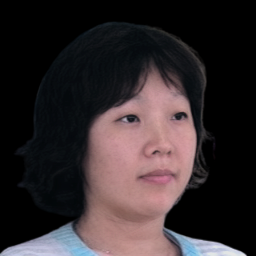}
    \caption{Ours}
\end{subfigure}
\begin{subfigure}[t]{\imgCrop\textwidth}
    \includegraphics[height=\textwidth,width=\textwidth]{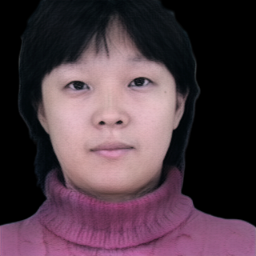}
    \includegraphics[height=\textwidth,width=\textwidth]{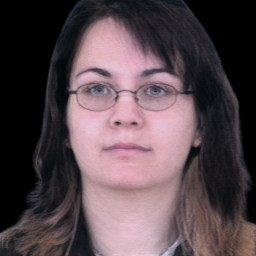}
    \includegraphics[height=\textwidth,width=\textwidth]{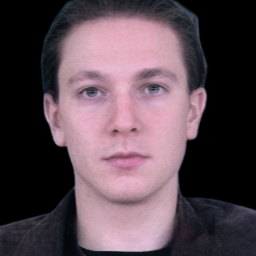}
    \includegraphics[height=\textwidth,width=\textwidth]{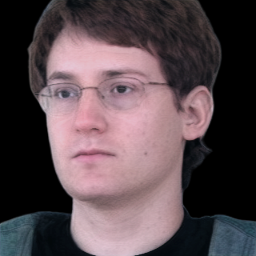}
    \includegraphics[height=\textwidth,width=\textwidth]{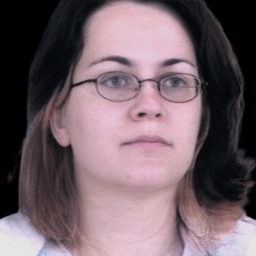}
    \includegraphics[height=\textwidth,width=\textwidth]{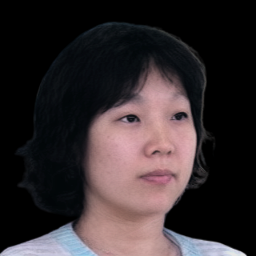}
    \caption{\centering TR (retrained)}
\end{subfigure}
\begin{subfigure}[t]{\imgCrop\textwidth}
    \includegraphics[height=\textwidth,width=\textwidth]{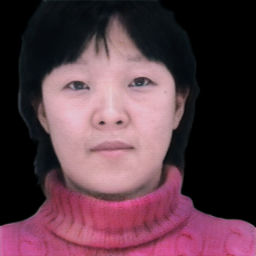}
    \includegraphics[height=\textwidth,width=\textwidth]{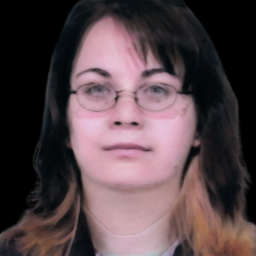}
    \includegraphics[height=\textwidth,width=\textwidth]{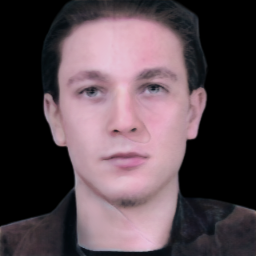}
    \includegraphics[height=\textwidth,width=\textwidth]{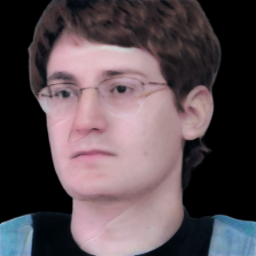}
    \includegraphics[height=\textwidth,width=\textwidth]{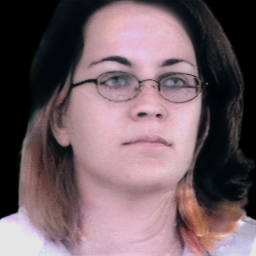}
    \includegraphics[height=\textwidth,width=\textwidth]{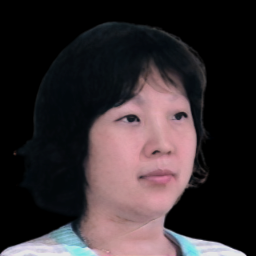}
    \caption{\centering EMR (retrained)}
\end{subfigure}
\begin{subfigure}[t]{\imgCrop\textwidth}
    \includegraphics[height=\textwidth,width=\textwidth]{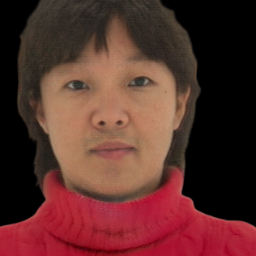}
    \includegraphics[height=\textwidth,width=\textwidth]{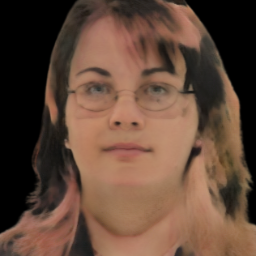}
    \includegraphics[height=\textwidth,width=\textwidth]{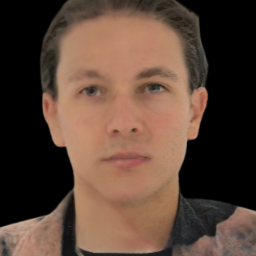}
    \includegraphics[height=\textwidth,width=\textwidth]{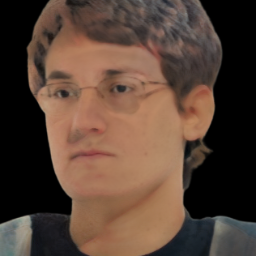}
    \includegraphics[height=\textwidth,width=\textwidth]{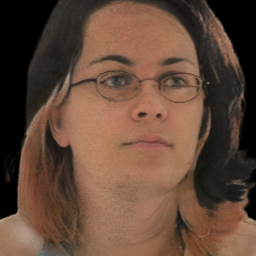}
    \includegraphics[height=\textwidth,width=\textwidth]{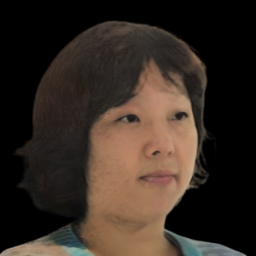}
    \caption{\centering TR}
\end{subfigure}
\begin{subfigure}[t]{\imgCrop\textwidth}
    \includegraphics[height=\textwidth,width=\textwidth]{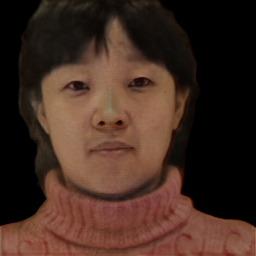}
    \includegraphics[height=\textwidth,width=\textwidth]{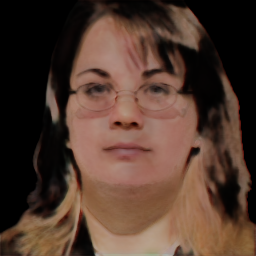}
    \includegraphics[height=\textwidth,width=\textwidth]{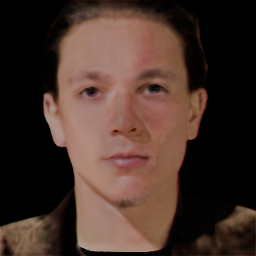}
    \includegraphics[height=\textwidth,width=\textwidth]{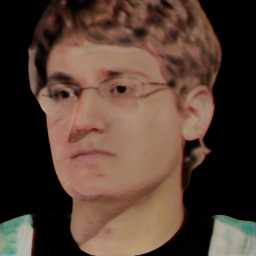}
    \includegraphics[height=\textwidth,width=\textwidth]{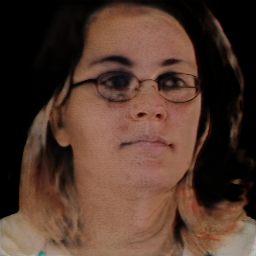}
    \includegraphics[height=\textwidth,width=\textwidth]{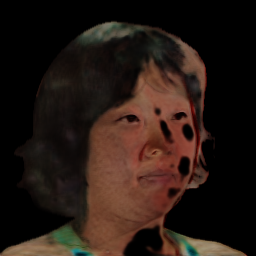}
    \caption{\centering EMR}
\end{subfigure}
\caption{We show more comparisons against other methods using face cropped images from our testing dataset.}
\label{face-crop-comparison}
\end{figure*}

\newcommand{\softScale}{0.21}
\begin{figure*}
\centering
\begin{subfigure}[t]{\softScale\linewidth}
    \includegraphics[height=\textwidth,width=\textwidth]{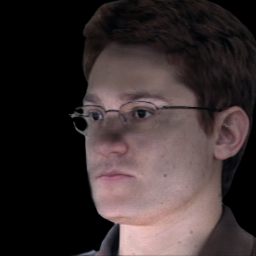}
    \includegraphics[height=\textwidth,width=\textwidth]{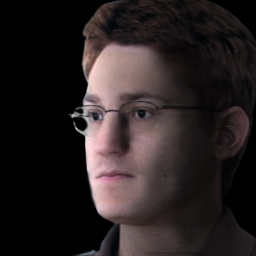}
    \includegraphics[height=\textwidth,width=\textwidth]{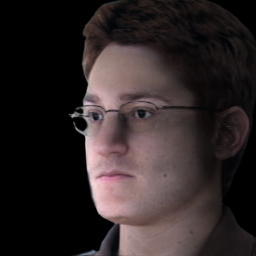}
    \includegraphics[height=\textwidth,width=\textwidth]{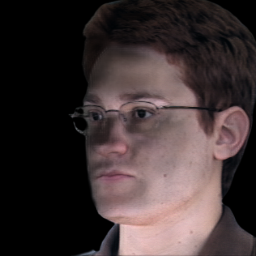}
    \includegraphics[height=\textwidth,width=\textwidth]{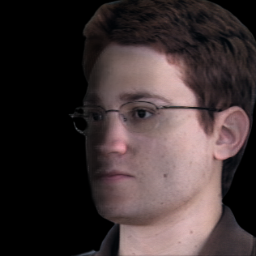}
    \includegraphics[height=\textwidth,width=\textwidth]{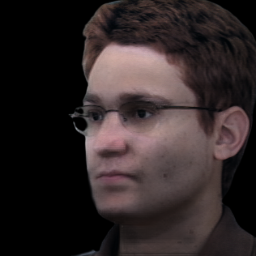}
    \caption{\centering Input}
\end{subfigure}
\begin{subfigure}[t]{\softScale\linewidth}
    \includegraphics[height=\textwidth,width=\textwidth]{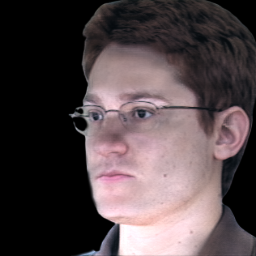}
    \includegraphics[height=\textwidth,width=\textwidth]{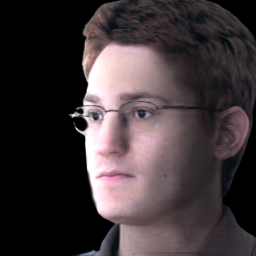}
    \includegraphics[height=\textwidth,width=\textwidth]{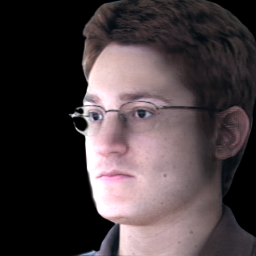}
    \includegraphics[height=\textwidth,width=\textwidth]{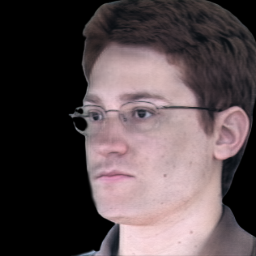}
    \includegraphics[height=\textwidth,width=\textwidth]{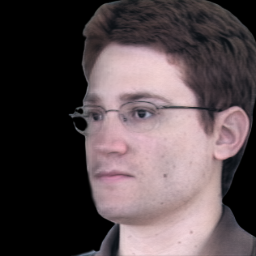}
    \includegraphics[height=\textwidth,width=\textwidth]{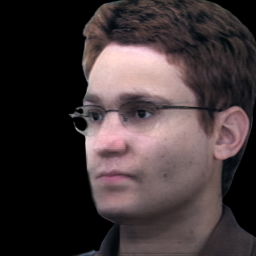}
    \caption{\centering w/o soft-shadow}
\end{subfigure}
\begin{subfigure}[t]{\softScale\linewidth}
    \includegraphics[height=\textwidth,width=\textwidth]{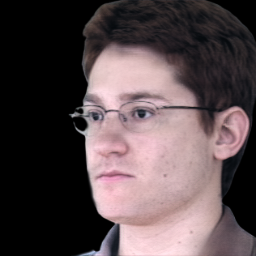}
    \includegraphics[height=\textwidth,width=\textwidth]{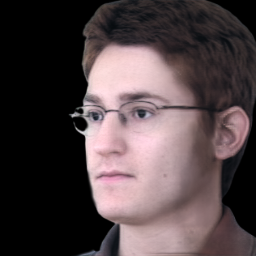}
    \includegraphics[height=\textwidth,width=\textwidth]{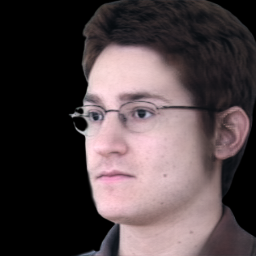}
    \includegraphics[height=\textwidth,width=\textwidth]{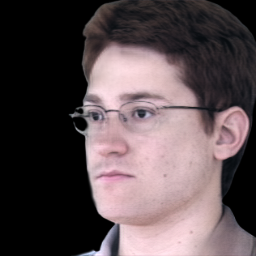}
    \includegraphics[height=\textwidth,width=\textwidth]{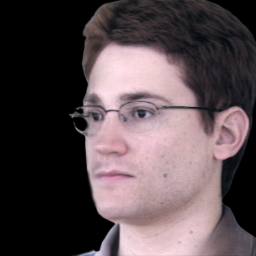}
    \includegraphics[height=\textwidth,width=\textwidth]{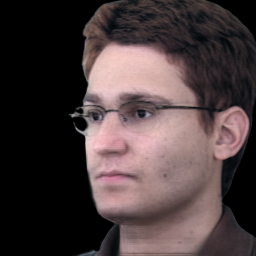}
    \caption{\centering w/ soft-shadow}
\end{subfigure}
\begin{subfigure}[t]{\softScale\linewidth}
    \includegraphics[height=\textwidth,width=\textwidth]{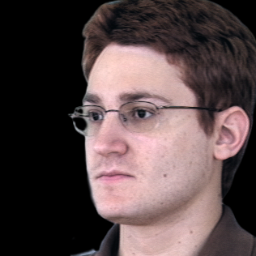}
    \includegraphics[height=\textwidth,width=\textwidth]{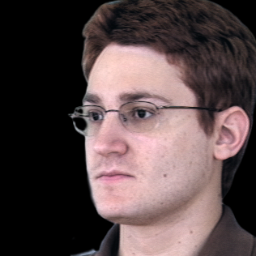}
    \includegraphics[height=\textwidth,width=\textwidth]{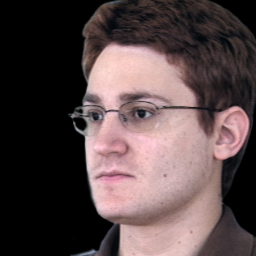}
    \includegraphics[height=\textwidth,width=\textwidth]{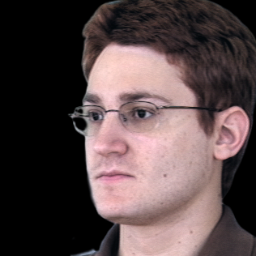}
    \includegraphics[height=\textwidth,width=\textwidth]{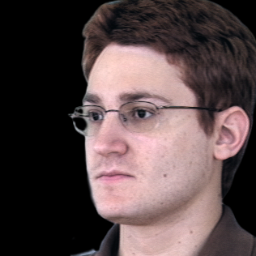}
    \includegraphics[height=\textwidth,width=\textwidth]{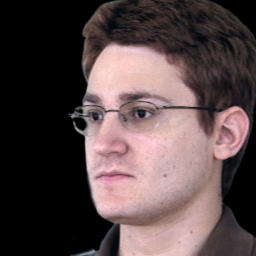}
    \caption{\centering Ground truth}
\end{subfigure}
\caption{\textbf{Soft-shadow ablation:} Input images from \textbf{Alt. Lighting} dataset. }
\label{soft-ablation}
\end{figure*}

\newcommand{\foreignScale}{0.195}
\begin{figure}
\centering
\begin{subfigure}[t]{\foreignScale\linewidth}
    \includegraphics[height=\textwidth,width=\textwidth]{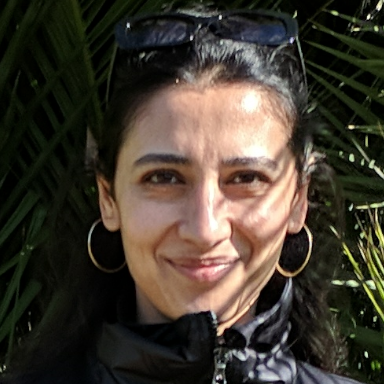}
    \includegraphics[height=\textwidth,width=\textwidth]{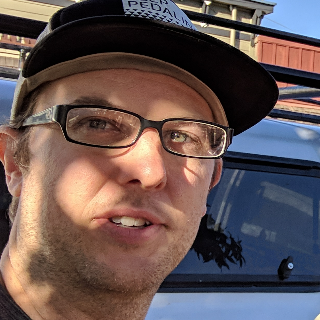}
     \includegraphics[height=\textwidth,width=\textwidth]{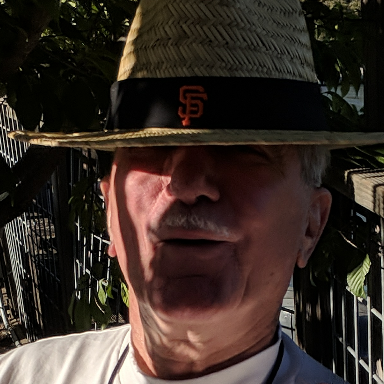}
    \includegraphics[height=\textwidth,width=\textwidth]{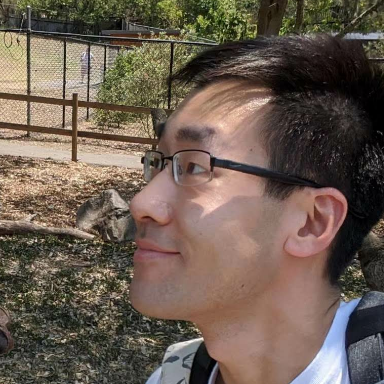}
    \includegraphics[height=\textwidth,width=\textwidth]{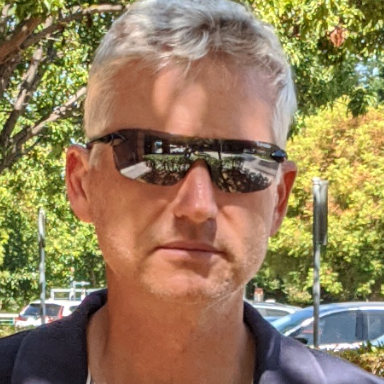}
    \includegraphics[height=\textwidth,width=\textwidth]{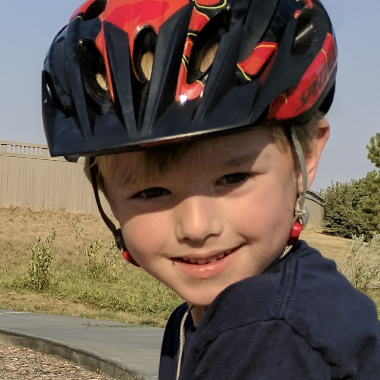}
    \includegraphics[height=\textwidth,width=\textwidth]{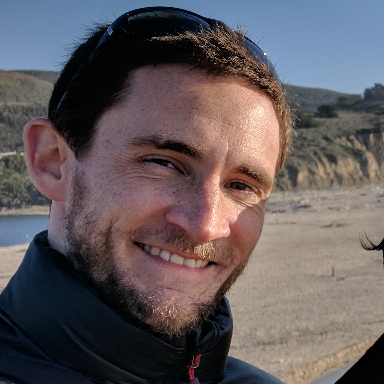}
    \caption{\centering Input}
\end{subfigure}
\begin{subfigure}[t]{\foreignScale\linewidth}
    \includegraphics[height=\textwidth,width=\textwidth]{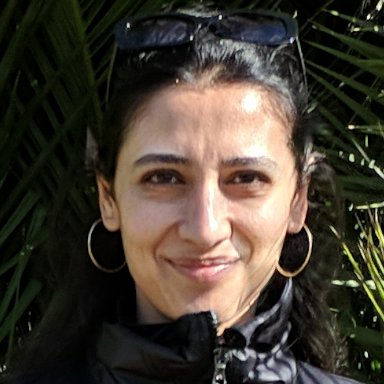}
    \includegraphics[height=\textwidth,width=\textwidth]{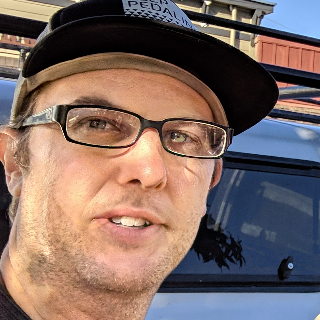}
    \includegraphics[height=\textwidth,width=\textwidth]{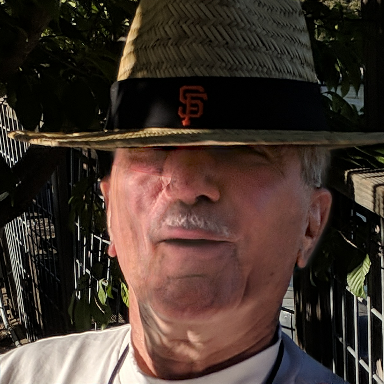}
    \includegraphics[height=\textwidth,width=\textwidth]{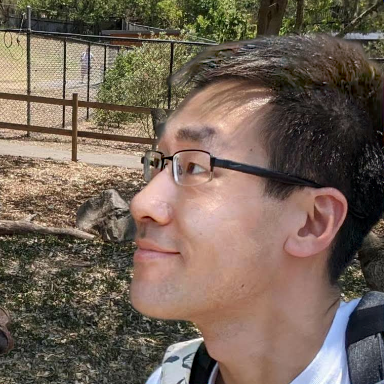}
    \includegraphics[height=\textwidth,width=\textwidth]{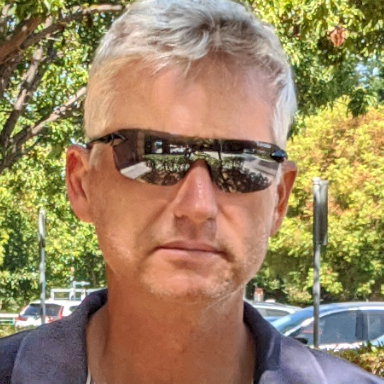}
    \includegraphics[height=\textwidth,width=\textwidth]{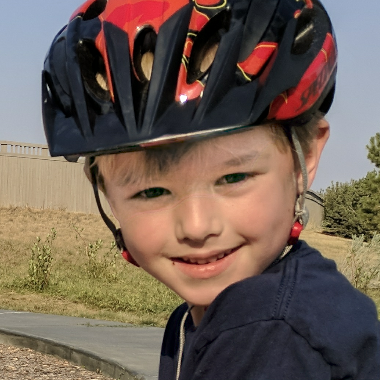}
    \includegraphics[height=\textwidth,width=\textwidth]{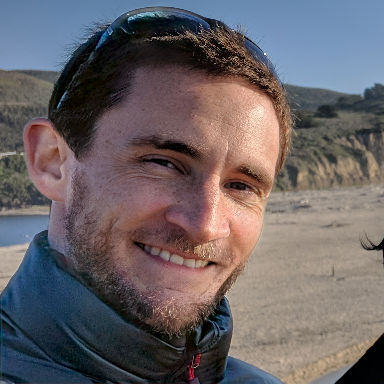}
    \caption{\centering Zhang \etal \cite{zhang2020portrait}}
\end{subfigure}
\begin{subfigure}[t]{\foreignScale\linewidth}
    \includegraphics[height=\textwidth,width=\textwidth]{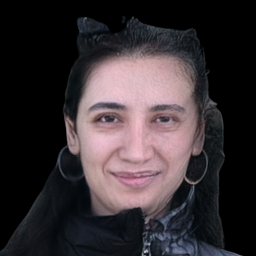}
    \includegraphics[height=\textwidth,width=\textwidth]{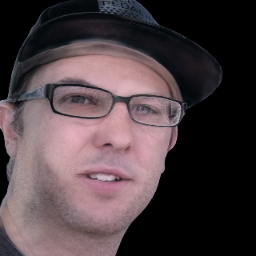}
    \includegraphics[height=\textwidth,width=\textwidth]{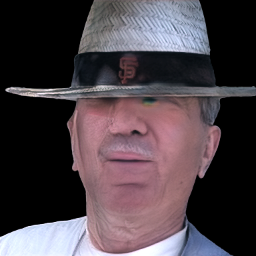}
    \includegraphics[height=\textwidth,width=\textwidth]{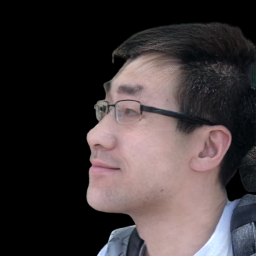}
    \includegraphics[height=\textwidth,width=\textwidth]{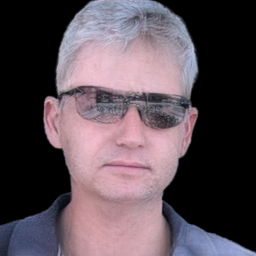}
    \includegraphics[height=\textwidth,width=\textwidth]{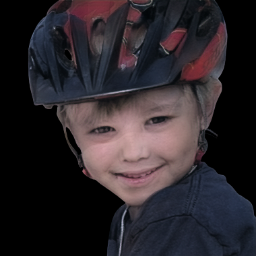}
    \includegraphics[height=\textwidth,width=\textwidth]{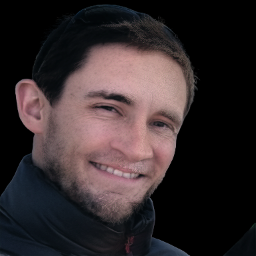}
    \caption{\centering Ours}
\end{subfigure}
\caption{\textbf{Shadow Removal Comparison 1:} Comparison with the method of Zhang \etal \cite{zhang2020portrait} on irregularly shaped shadows. All images in (a) and (b) are taken from the \href{https://ceciliavision.github.io/project-pages/portrait}{author's website.}}
\label{shadow_removal_2}
\end{figure}